\documentclass[sigconf]{acmart}
\AtBeginDocument{%
  }

\setcopyright{none}
\copyrightyear{2025}
\acmYear{2025}
\acmDOI{}

\acmConference[LLM4ECommerce Workshop at KDD '25]{LLM4ECommerce Workshop at the 31st ACM SIGKDD Conference on Knowledge Discovery and Data Mining V.2}{August 4, 2025}{Toronto, ON, Canada}
\acmISBN{}

\usepackage[utf8]{inputenc} 
\usepackage[T1]{fontenc}    
\usepackage{hyperref}       
\usepackage{url}            
\usepackage{booktabs}       
\usepackage{amsfonts}       
\usepackage{nicefrac}       
\usepackage{microtype}      
\usepackage{xcolor}  

\usepackage{pdfpages}
\usepackage[absolute]{textpos}
\usepackage{microtype}
\usepackage{graphicx}
\usepackage{algorithm}
\usepackage{algpseudocode}
\usepackage{amsfonts}
\usepackage{transparent}
\usepackage{subfigure}
\usepackage{booktabs} 
\usepackage{enumitem}
\usepackage{xspace}
\usepackage{xcolor}
\usepackage{soul}
\usepackage[most]{tcolorbox}

\definecolor{cyan10}{HTML}{E5F6FF}
\definecolor{cyan20}{HTML}{BAE6FF}
\definecolor{cyan60}{HTML}{0072c3}
\definecolor{cyan70}{HTML}{00539a}
\definecolor{teal10}{HTML}{D9FBFB}
\definecolor{teal20}{HTML}{9EF0F0}
\definecolor{teal60}{HTML}{007d79}
\definecolor{orange10}{HTML}{FFF2E8}
\definecolor{orange20}{HTML}{FFD9BE}
\definecolor{orange60}{HTML}{ba4e00}
\definecolor{blue10}{HTML}{EDF5FF}
\definecolor{blue20}{HTML}{D0E2FF}
\definecolor{magenta10}{HTML}{FFF0F7}
\definecolor{magenta20}{HTML}{FFD6E8}
\definecolor{magenta30}{HTML}{ffafd2}
\definecolor{magenta50}{HTML}{ee5396}
\definecolor{magenta60}{HTML}{d02670}
\definecolor{magenta70}{HTML}{9f1853}
\definecolor{purple10}{HTML}{F6F2FF}
\definecolor{purple20}{HTML}{E8DAFF}
\definecolor{purple30}{HTML}{d4bbff}
\definecolor{purple70}{HTML}{8a3ffc}
\definecolor{rose10}{HTML}{FCF2ED}
\definecolor{rose20}{HTML}{F9D9D1}
\definecolor{red10}{HTML}{FFF1F1}
\definecolor{red20}{HTML}{FFD7D9}
\definecolor{green10}{HTML}{DEFBE6}
\definecolor{green20}{HTML}{A7F0BA}
\definecolor{yellow10}{HTML}{fcf4d6}
\definecolor{yellow20}{HTML}{fddc69}
\definecolor{gray20}{HTML}{e0e0e0}
\definecolor{gray30}{HTML}{c6c6c6}
\definecolor{gray40}{HTML}{a8a8a8}
\definecolor{gray80}{HTML}{393939}

\definecolor{carbon-gray-10}{cmyk}{0.0, 0.0, 0.0, 0.04, 1.00}
\definecolor{carbon-gray-90}{cmyk}{0.0, 0.0, 0.0, 0.85, 1.00}

\usepackage{listingsutf8}
\newcommand{\margen}{\textsc{MaRGen}\xspace}
\newcommand{\researcher}{\textsc{Researcher}\xspace}
\newcommand{\writer}{\textsc{Writer}\xspace}
\newcommand{\judge}{\textsc{Judge}\xspace}
\newcommand{\reviewer}{\textsc{Reviewer}\xspace}
\newcommand{\retriever}{\textsc{Retriever}\xspace}

\usepackage{amsmath}
\usepackage{mathtools}
\usepackage{amsthm}

\usepackage[capitalize,noabbrev]{cleveref}

\theoremstyle{plain}

\theoremstyle{definition}

\theoremstyle{remark}

\usepackage[textsize=tiny]{todonotes}

\title[\textsc{MaRGen}: Multi-Agent Self-Directed Market Research]{\textsc{MaRGen}: Multi-Agent LLM Approach for Self-Directed Market Research and Analysis}
\author{Roman Koshkin}
\authornote{Both authors contributed equally to this research.}
\affiliation{%
  \institution{Okinawa Institute of Science and Technology}
  \city{Onna}
  \state{Okinawa}
  \country{Japan}
}
\email{roman.koshkin@gmail.com}

\author{Pengyu Dai}
\authornotemark[1]
\affiliation{%
  \institution{Institute of Science Tokyo}
  \city{Tokyo}
  \country{Japan}}
\email{dai.p.aa@m.titech.ac.jp}

\author{Nozomi Fujikawa}
\affiliation{%
  \institution{Amazon}
  \city{Tokyo}
  \country{Japan}
}
\email{nozofuji@amazon.co.jp}

\author{Masahito Togami}
\affiliation{%
 \institution{Amazon}
 \city{Tokyo}
 \country{Japan}}
\email{mtogami@amazon.co.jp}

\author{Marco Visentini-Scarzanella}
\affiliation{%
  \institution{Amazon}
  \city{Tokyo}
  \country{Japan}}
\email{marcovs@amazon.co.jp}

\begin{document}
\begin{abstract}
We present an autonomous framework that leverages Large Language Models (LLMs) to automate end-to-end business analysis and market report generation. 
At its core, the system employs specialized agents - Researcher, Reviewer, Writer, and Retriever - that collaborate to analyze data and produce comprehensive reports. 
These agents learn from real professional consultants' presentation materials at Amazon through in-context learning to replicate professional analytical methodologies. 
The framework executes a multi-step process: querying databases, analyzing data, generating insights, creating visualizations, and composing market reports. 
We also introduce a novel LLM-based evaluation system for assessing report quality, which shows alignment with expert human evaluations.
Building on these evaluations, we implement an iterative improvement mechanism that optimizes report quality through automated review cycles. 
Experimental results show that report quality can be improved by both automated review cycles and consultants' unstructured knowledge.
In experimental validation, our framework generates detailed 6-page reports in 7 minutes at a cost of approximately \$1. 
Our work could be an important step to automatically create affordable market insights.
\end{abstract}

\ccsdesc[500]{Computing methodologies~Intelligent agents}
\ccsdesc[500]{Computing methodologies~Multi-agent systems}
\ccsdesc[500]{Computing methodologies~Natural language generation}
\ccsdesc[500]{Computing methodologies~Artificial intelligence}

\keywords{LLM, Multi-agent, report generation}
\begin{teaserfigure}
   \includegraphics[width=0.99\textwidth]{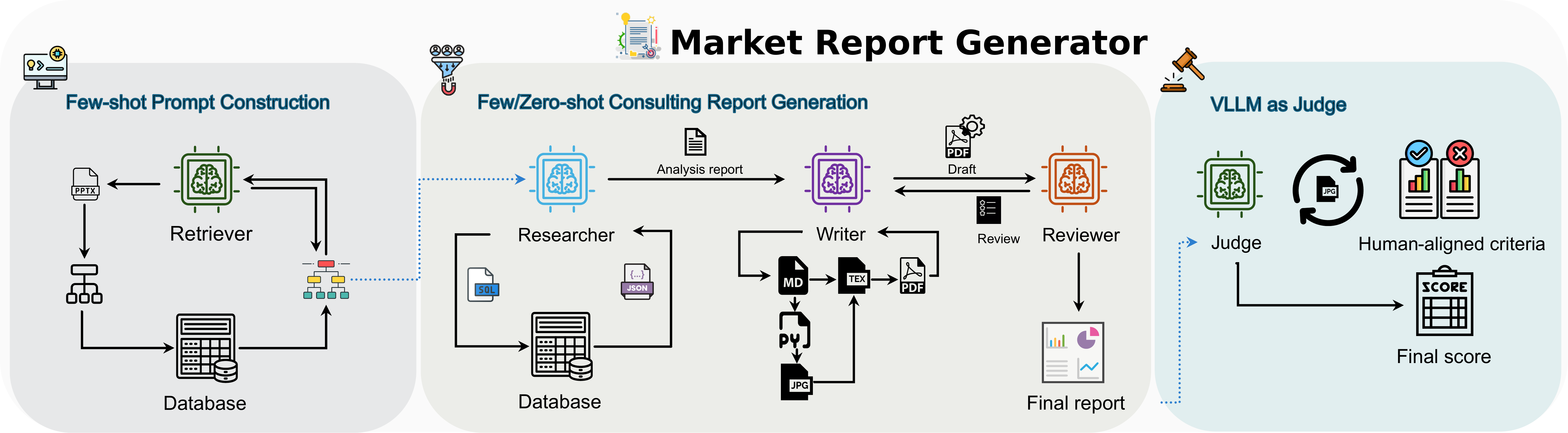}
   \caption{Overview of \margen. The \researcher writes a sequence of SQL queries (one at a time), each of which is executed and the returned data are added to the prompt. This process continues until the LLM decides that it has enough information to write a research report or exceeds \texttt{MAX\_QUERIES}. Based on the research report, the \writer creates relevant figures and LaTeX code, which are compiled to PDF and sent to the \reviewer, which suggests improvements. After several rounds of review, the final report is returned to the user. \margen generates multiple reports and only the best of them (as decided by the \judge agent) will be returned to the user. Additionally, the \retriever agent can extract business insights from available unstructured materials, which can be fed to the \researcher and help improve the quality of the final report.}
   \label{fig:fig2}
\end{teaserfigure}

\maketitle

\section{Introduction}
\label{intro}
In today's fast-paced e-commerce environment, 
the extraction of actionable insights from vast and diverse datasets encompassing customer 
behavior, product performance, and related variables - 
and the subsequent application of these insights to strategic decision-making processes - 
represents a critical imperative for contemporary organizations.
However, many organizations face significant challenges in efficiently analyzing data due 
to difficulty of securing skilled business / data analysts.

To resolve this challenge and streamline data-driven decision making through automated e-commerce analytics and insight generation, we introduce \textsc{Market Report Generator} (\margen), 
a comprehensive and fully automated framework designed to accelerate market research. 
Fig.~\ref{fig:fig2} offers an overview of \margen. 
\margen implements a multi-agent system that automates business analysis—from interpreting client requirements to querying databases and generating analytical reports with supporting visualizations. \margen distills domain expertise and reasoning from unstructured text authored by professional human consultants and incorporates by in-context learning.
For real-world applications, evaluation of report quality is essential. 
We further propose an automated evaluation framework based on pairwise comparisons of candidate reports, which achieved a Pearson correlation of 0.6 ($p < 0.01$) with expert human evaluations. 

Our main contributions are as follows:

\begin{enumerate}
\item We propose a method to extract Amazon professional consultants' expertise from unstructured materials (PowerPoint) and exploit it via few-shot prompting, demonstrating that this \textbf{domain knowledge} significantly improves report quality.
\item We propose a \textbf{versatile multi-agent framework}  to automate market analysis. Given high-level request, it produces comprehensive reports without requiring domain expertise, using a collaborative multi-agent process for tasks like SQL querying, data analysis, and report generation.
\item We propose an \textbf{automated report evaluation framework} of LLMs, showing alignment with human assessments. Based on this, we further propose a LLM-based review mechanism and an approach for selecting optimal reports, enabling \textbf{autonomous and iterative improvement} in report generation.
\end{enumerate}

\section{Related Work}
\label{sec:relatedwork}
While Large Language Models (LLMs) have demonstrated remarkable capabilities, certain conventional architectures—particularly those based on token-level, left-to-right sequence generation—present challenges in complex problem solving that require extensive exploration and reasoning.
Researchers have addressed this by embedding structured reasoning processes that mirror human cognitive patterns.
Chain-of-Thought (CoT) prompting~\citep{wei2022chain} emerged as an early influential research in 
this field, followed by Tree of Thoughts (ToT)~\citep{yao2023tree},
Graph of Thoughts (GoT)~\citep{besta2024graph}, and Buffer of Thoughts (BoT)~\citep{yang2024buffer}. 
These approaches typically decompose complex reasoning into manageable units 
—such as individual steps of logical reasoning that might generate output of appropriate granularity— 
that are neither too large nor too small for effective LLM processing, 
integrating these units into comprehensive problem-solving frameworks~\citep{chen2025towards}.
Recent research has also explored ways to augment LLM capabilities and address limitations such as 
lack of persistent memory, outdated knowledge, and hallucination tendencies~\citep{minaee2025largelanguagemodelssurvey}. 
Solutions include Retrieval-Augmented Generation (RAG)~\citep{lewis2020retrieval,gao2023retrieval} for accessing latest information, 
external tool (e.g., an API to a service) integration~\citep{schick2023toolformerlanguagemodelsteach,paranjape2023artautomaticmultistepreasoning}, 
and agentic approach architected to perform complex operations requiring 
decision-making and action~\citep{guo2024large,xi2025rise,Wang_2024,durante2024agentaisurveyinghorizons}. 
Our research investigates a novel application in this evolving landscape: the automation of market 
analysis through iterative data exploration with SQL, insight extraction, 
and comprehensive report creation with visualizations.

Recent research on LLMs for complex data analysis can be categorized into two types: 
problems with ground-truth solutions and those without objective solutions. 
Zhang et al.~\citep{coddllm} presents \textit{CODDLLM}, 
a post-trained LLM for data analytics tasks like database table selection, addressing the first category.
For tasks without objective solutions, two complementary approaches emerge: 
combining non-LLM analytical methods with LLMs, 
and allowing LLMs to generate code dynamically. 
Ma et al.~\citep{ma-etal-2023-insightpilot} demonstrates the first approach 
with \textit{InsightPilot}, using LLMs to guide insight generation with 
predefined functions (such as trend prediction algorithm) that have been carefully designed. 
While this reduces hallucinations, it has limitations in flexibility.
For the second approach, research explores multi-agent systems for complex tasks~\citep{lu2024aiscientistfullyautomated,gottweis2025aicoscientist}. 
P{'e}rez et al.~\citep{perez-etal-2025-llm} proposes a multi-agent system consisting of a \textit{Hypothesis Generator} that creates high-level questions, a \textit{Query Agent} that 
validates them using SQL queries, and a \textit{Summarization} module that consolidates the results into insights.

Our research, while also utilizing a multi-agent system, uniquely leverages domain knowledge 
from professional business consultants.

\section{Methods}
\label{sec:methods}

In this section, we formalize \margen, the proposed multi-agent framework designed for automatic business analysis. Fig.~\ref{fig:fig2} offers an overview of \margen. Given a complex consulting task \(x\), the framework generates a business analysis report \(y\) by decomposing the analysis process into multiple subtasks, each handled by specialized agents. The complete \margen framework is summarized in Algorithm \ref{alg:symbolic_agents} and elaborated on below. \margen consists of four agents:  \researcher, \writer, \reviewer, and \retriever, each of which will be introduced in detail below.

\small 

\begin{algorithm}[!ht]
\caption{MaRGen Framework}
\label{alg:symbolic_agents}
\begin{algorithmic}[1]
\Statex \textbf{Initialize:}
\State $N \leftarrow \text{MaxRounds}$ \Comment{Max iterations}
\State $t \leftarrow 0$ \Comment{Current index}
\State $\mathcal{K} \leftarrow \emptyset$ \Comment{External knowledge}

\If{Few-Shot}
    \State $\mathcal{K} \leftarrow \text{Retriever}(x)$ \Comment{Retrieve knowledge}
\EndIf

\Statex \textbf{Researcher:}
\If{Few-Shot}
    \State $o_0 \leftarrow p_{\theta}(o \mid I, \mathcal{K}, x)$ \Comment{Init with knowledge}
\Else
    \State $o_0 \leftarrow p_{\theta}(o \mid I, x)$ \Comment{Random init}
\EndIf
\State $q_0 \leftarrow E(o_0)$; \quad $d_0 \leftarrow \text{DB}(q_0)$

\While{$t < N$ \textbf{and} $\text{NeedMoreData}(o_{t}, d_{t})$}
    \State $o_t \leftarrow p_{\theta}(o \mid I, \{(o_i, d_i)\}_{i=0}^{t-1})$
    \State $q_t \leftarrow E(o_t)$; \quad $d_t \leftarrow \text{DB}(q_t)$
    \State $t \leftarrow t + 1$
\EndWhile
\State $H \leftarrow \{(o_i, d_i)\}_{i=0}^{t-1} \cup \{o_T\}$ \Comment{Results}
\State \textbf{Researcher} $\Rightarrow$ \textbf{Writer}: $H$

\Statex \textbf{Writer:}
\State $R_t^{\text{md}} \leftarrow p_{\theta}(\cdot \mid H, I_{\text{wr}})$ \Comment{Markdown}
\State $(p_1, \ldots, p_k) \leftarrow \text{Extract\_Code}(R_t^{\text{md}})$
\State $\text{Execute}(p_1, \ldots, p_k)$ \Comment{Run code}
\State $R_t^{\text{tex}} \leftarrow \text{Convert\_To\_LaTeX}(R_t^{\text{md}})$
\State $R_t^{\text{pdf}} \leftarrow \text{Compile\_PDF}(R_t^{\text{tex}})$
\State \textbf{Writer} $\Rightarrow$ \textbf{Reviewer}: $R_t^{\text{pdf}}$

\Statex \textbf{Reviewer:}
\State $F_t \leftarrow p_{\theta}(F \mid R_t^{\text{pdf}}, I_{\text{rev}})$ \Comment{Evaluate}
\If{$F_t < \text{Threshold}$}
    \State $R_t^{\text{md}} \leftarrow \text{RequestRevision}(F_t, R_t^{\text{pdf}})$
    \State \textbf{Reviewer} $\Rightarrow$ \textbf{Writer}: $F_t$
\Else
    \State \textbf{Approve report}
\EndIf

\Statex \textbf{Iteration \& Termination:}
\While{$t < N$}
    \State \textbf{Writer}: $R_t^{\text{md}} \rightarrow R_t^{\text{pdf}}$
    \State \textbf{Reviewer}: Evaluate $F_t$
    \If{$F_t \geq \text{Threshold}$}
        \State \textbf{Terminate}
    \EndIf
    \State $t \leftarrow t + 1$
\EndWhile
\State \Return $y \leftarrow R_t^{\text{pdf}}$ \Comment{Final report}
\end{algorithmic}
\end{algorithm}

\normalsize

\subsection{Researcher} 

Motivated by the hypothetical principles of human decision-making and game theory~\cite{simon1957models}, given a prompt including a minimal description of the client and their objectives, the \researcher proposes a initial hypothesis, gathers data to back it up, interprets and summarizes the findings. Its initial prompt (see Appendix \ref{appendix:prompts}) may include general information about the client and their objectives, as well as constraints to limit the research space. For example, we might want to focus on a particular period (to reduce the number of rows to scan in the database), limit the number of SQL queries to perform, and rows returned by one query etc. Further, the prompt should include information about the expected formatting of the output (e.g. instruction to surround the final report by the opening and closing tags \texttt{<FINAL\_ANSWER>} and \texttt{</FINAL\_ANSWER>} to facilitate its downstream processing (see Appendix \ref{appendix:prompts}). Given this prompt, the LLM generates an answer, from which we parse an SQL query, execute it and append the result to the next prompt for the LLM to interpret. This process continues until the LLM either has enough data to provide the final answer or the maximum number of queries is reached.

The \researcher's full pipeline is as follows:

\begin{enumerate}
    \item Propose an initial hypothesis, decide what data are necessary to support the hypothesis, write the first SQL query;
    \item Execute the query; 
    \item Interpret the result of the query, and update the initial hypothesis as necessary
        \begin{itemize}
            \item if the number of queries is less than \texttt{MAX\_QUERIES}, and if more data needs to be collected, write the next query and go to Step 2;
            \item otherwise summarize the data collected so far and return the summary
        \end{itemize}
\end{enumerate}

Formally, the \researcher's pipeline is defined in Eq. \ref{eq:1}:

\begin{equation}
o_t \sim p_{\theta}\big(\cdot|I, \{o_{i}, d_{i}\}_{i=0}^{t-1}\big)
\label{eq:1}
\end{equation}

where $p_\theta$ is an LLM, $I$ is the initial prompt (including the system message), $d_i = \text{DB}(\text{E}(o_i))$ is the data returned by a SQL query to a database at the $i$-th time step, $\text{E}: o \rightarrow q$ is a script that extracts the SQL query, $q$, from the LLM's output, $o$, and $\text{DB}: q \rightarrow d$ is the database that returns data, $d$, given a query, $q$.

After the \researcher has finished its work, the entire research history is passed as part of the prompt to the \writer agent, which we describe next.

\subsection{Writer}

The \writer creates the first version of the final report in markdown format (for the full prompt refer to Appendix \ref{appendix:prompts}).

\begin{equation}
R_0^{\text{md}} \sim p_{\theta}\big(\cdot|I_{\text{wr}}^{\text{md}}, H\big)
\end{equation}

where $p_\theta$ is an LLM, $I^{\text{md}}_{\text{wr}}$ is the prompt, including the system message, to generate a report in markdown format, $H = \{o_{i}, d_{i}\}_{i=0}^{T-1} \oplus \{o_T\}$ is the full history of research produced by the \researcher. In subsequent rounds, the \writer generates reports conditioned on the feedback from the \reviewer and  its own reports generated previously:

\begin{equation}
R_t^{\text{md}} \sim p_{\theta}\big(\cdot|I_{\text{wr}}^{\text{md}}, H, \{R^{\text{md}}_{i}, F_{i}\}_{i=0}^{t-1}\big)
\end{equation}

where $R^{\text{md}}_i$ and $F_i$ are a previous markdown report and \reviewer's feedback, up to the previous step (inclusive). $R^{\text{md}}_i$ contains the text of the report and, in their appropriate places, blocks of python code for generating each figure. Before the \writer can generate latex code, it needs to extract the Python code blocks and run them to generate and save the corresponding figures:

\begin{equation}
(p_1, \dots, p_N) = \text{EF}(R^{\text{md}}_t)
\end{equation}

where $\{p\}_{i=0}^N$ are python blocks and $\text{EF}: R^{\text{md}} \rightarrow \{p\}_{i=1}^N$ is a script that extracts $N$ python blocks from the LLM's output (the markdown text in this case).

Next, the \writer generates latex code from the markdown document, replacing python code blocks with references to previously rendered figures, as follows:

\begin{equation}
R_t^{\text{tex}} \sim p_{\theta}\big(\cdot|I_{\text{wr}}^{\text{md}}, H, \{R^{\text{md}}_{i}, F_{i}\}_{i=0}^{t-1}, R^{\text{md}}_t\big)
\end{equation}

where $R^{\text{tex}}_t$ is latex code corresponding to $R^{\text{md}}_t$ and $I_{\text{wr}}^{\text{md}}$ is the prompt, which includes a concise style template, to convert it into latex code, which is rendered into a PDF document.

In all rounds except the first and last, before submitting the PDF rendered from $R^{\text{tex}}_t$ for review, the \writer additionally generates a response to the \reviewer, detailing what has been done specifically to address the \reviewer's concerns and comments:

\begin{equation}
A_t \sim p_{\theta}\big(\cdot|I_{\text{ans}}^{\text{md}}, H, \{R^{\text{md}}_{i}, F_{i}\}_{i=0}^{t-1}, R^{\text{pdf}}_t \big)
\end{equation}

Note that the LLM cannot directly process PDF documents, so we convert each PDF report into a sequence of images, $R^{\text{pdf}}_t$, one image (600 by 600 pixels) for each page of the PDF. On receiving the review (including the scores), the \writer implements the changes suggested by the \reviewer by generating an improved report. This process is repeated for several rounds or until either the draft gets the perfect score of 10 on both "clarity" and "layout", or a maximum number of rounds is reached. The \writer's pipeline can be summarized as follows:

\begin{enumerate}
    \item Generate a report in markdown format, implementing the edits suggested by the reviewer (except in Round 0), with blocks of Python code inserted in the appropriate sections;
    \item Extract the python code blocks, run them and save the figures as images files;
    \item Based on the markdown file, generate latex code, inserting references to the image files and compile to a PDF document;
    \item Send the PDF to the \reviewer;
    \item Receive the feedback and scores from the \reviewer; exit if perfect score or after 4 rounds, otherwise go back to Step 1.
\end{enumerate}

\subsection{Reviewer}

The \reviewer agent reviews and scores the PDF report on a scale of 1 to 10 (where 1 is poor and 10 is excellent) written by the \writer based on the following two criteria:  \emph{clarity} and \emph{layout} (the full prompt is provided in Appendix \ref{appendix:reviewers_protmps}) and returns the review and scores back to the \writer for generating an improved version of the report in the next round.

\begin{equation}
F_t \sim p_{\theta}\big(\cdot|I_{\text{rev}}, \{F_{i}, A_{i}\}_{i=0}^{t-1}, R^{\text{pdf}}_{t}\big)
\end{equation}

where $p_\theta$ is an LLM, $I_{\text{rev}}$ is the prompt, including the system message, to review the latest report $R^{\text{pdf}}_t$ given the full history (if any) of previous reviews and \writer's replies to the \reviewer, $\{F_{i}, A_{i}\}_{i=0}^{t-1}$, up to the previous round (inclusive). Note that in order to allow the \reviewer to evaluate the overall appearance of the document (e.g. element layout, font and color choices), the latest PDF report is fed to the LLM in an image format.

\subsection{Retriever}
\label{subsec:Retriever}
The content produced by the \researcher agent is based solely on historical data and its own 
(largely unconstrained) reasoning about these data. 
However, we could improve report quality by aligning the agent's reasoning with human domain experts.
To achieve this, we propose a \retriever agent that extracts examples of professional reasoning 
from Amazon consultants' materials (PowerPoint). 
These examples are then inserted as few-shot examples in the \researcher's prompt.

\begin{figure}[h!]
   \centering
   \includegraphics[width=0.99\columnwidth]{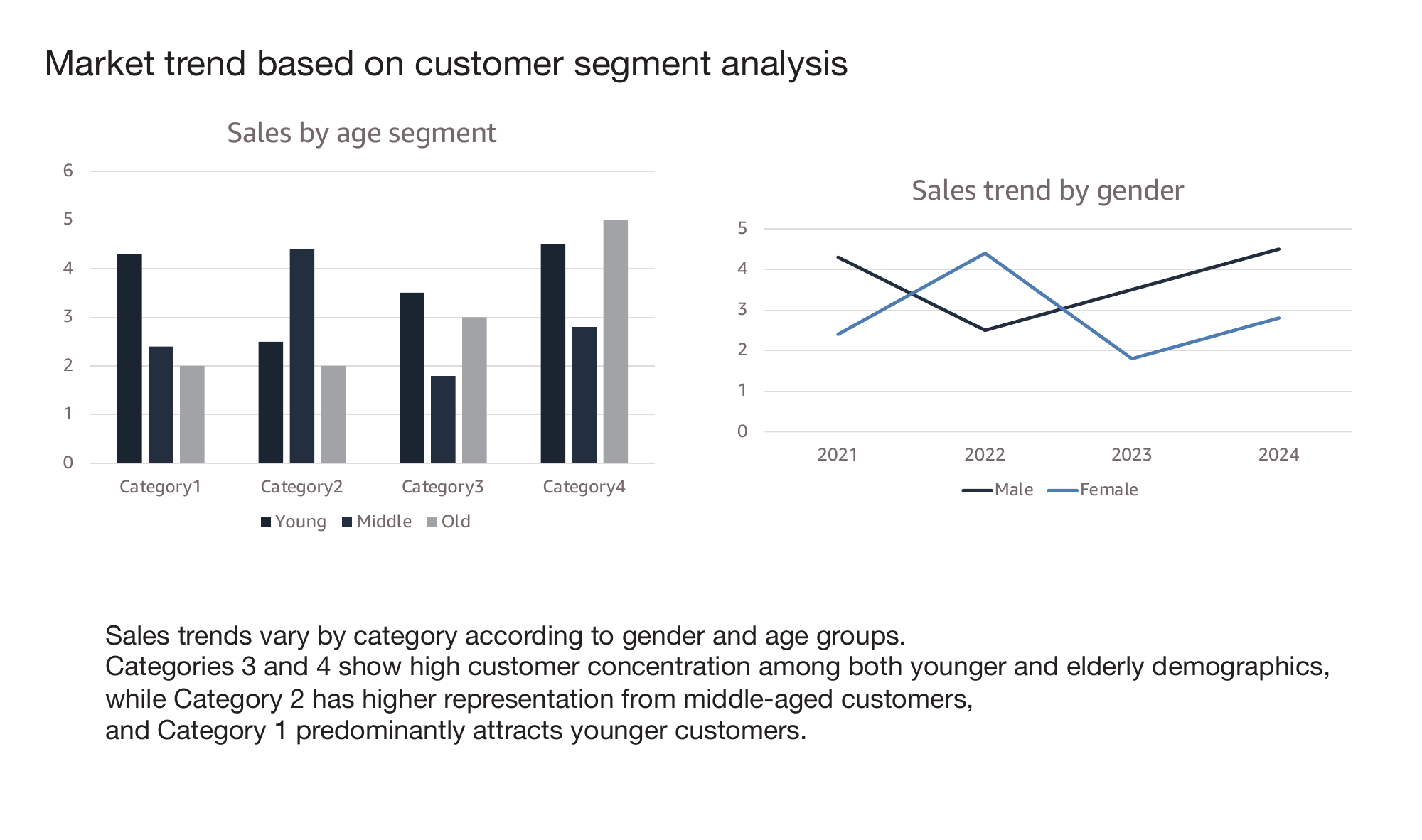}
   \caption{\label{fig:sample_ppt} Example of a conceptual consultant's PowerPoint slide.}
\end{figure}

\begin{figure}[h!]
   \centering
   \includegraphics[width=0.99\columnwidth]{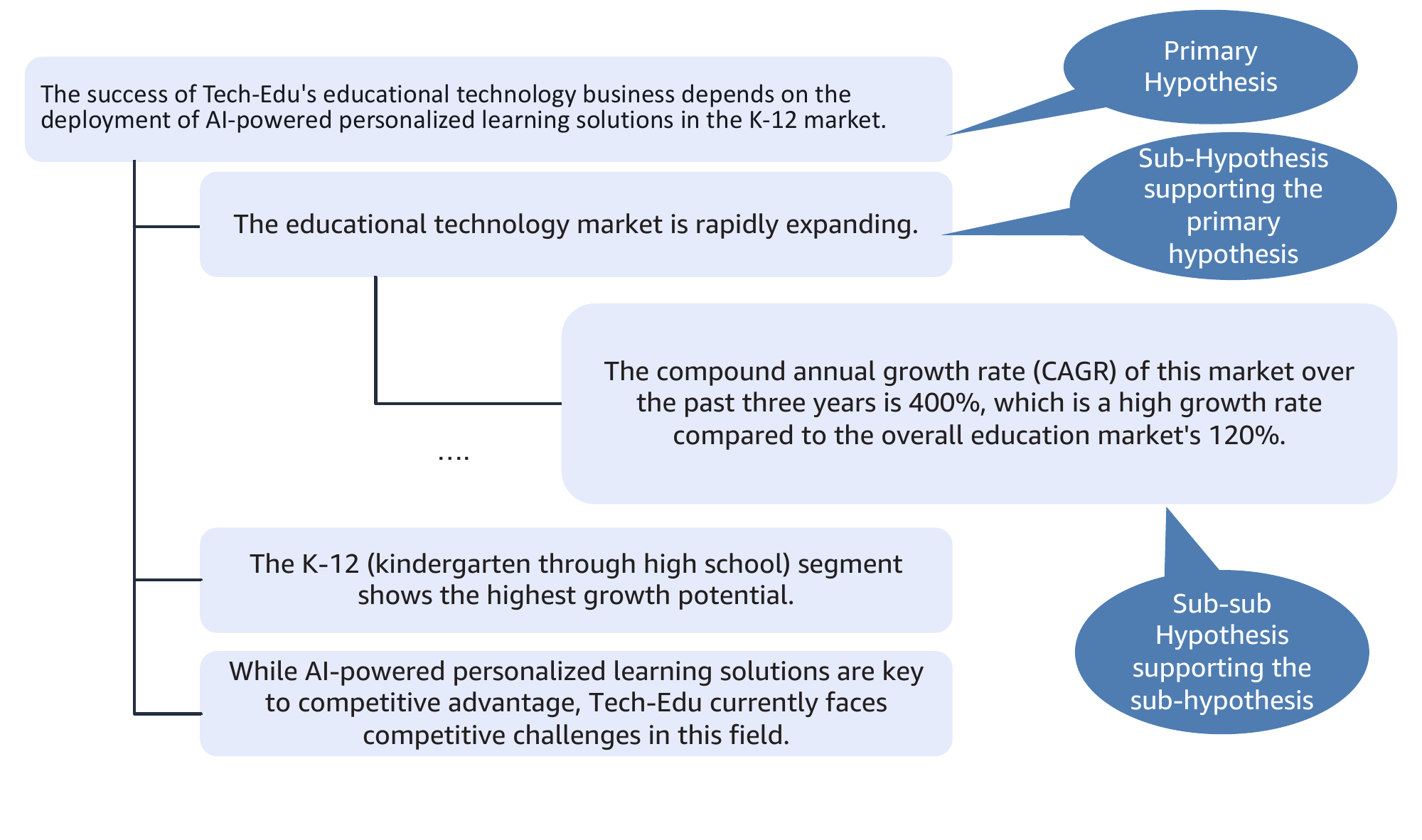}
   \caption{\label{fig:ppt_tree} Example of a conceptual hypothesis tree structure extracted from an actual consultant's PowerPoint slide.}
\end{figure}

Among professional business consultants' methodologies, 
Barbara Minto's Pyramid Principle \citep{minto2009pyramid} is a key framework. 
This approach places the conclusion (hypothesis) at the top, 
with supporting evidence arranged hierarchically below for logical coherence. 
Such hierarchical relationships can be represented as a tree structure, 
which typically has a limited depth. Thus, from consulting materials, 
we extract consultants' domain knowledge by deriving tree structures with a depth of 2. 
For a conceptual example of consultants' PowerPoint slide, see Figure~\ref{fig:sample_ppt}.
For a conceptual example tree derived from consultants' materials, see Figure~\ref{fig:ppt_tree}.
Please note that the original PowerPoint material created by consultants consists of multiple slides containing various figures and texts, and is an unstructured document. \retriever agent extracts hypothesis trees like the one shown in Figure~\ref{fig:ppt_tree} from this unstructured data, inserting them as few-shot examples in the \researcher's prompt.

Specifically, the root node (primary hypothesis) represents high-level strategic decisions or judgments, 
the middle nodes (sub-hypotheses) represent supporting arguments that justify the root node, 
and the leaf nodes (sub-sub hypotheses) represent specific data-based evidence supporting the middle nodes.

Formally, assume that for a given client there is a collection of $n$ documents \(\mathcal{D} = \{d_1, d_2, \dots, d_n\}\), where each \(d_i\) represents a multimodal document containing both text and images. We first utilize a multi-modal embedding model to construct corresponding retrieval vectors $\{\mathbf{v}_i\}$:

\begin{equation}
\mathcal{K} = \{\mathbf{v}_i \sim \text{Em}(d_i) \mid d_i \in \mathcal{D}\},
\end{equation}

We leverage multi-modal conversations to extract hypotheses from documents, represented as $\mathcal{H} = \{h_1, h_2, \dots, h_m \}
$, where each $h_i$ denotes a distinct hypothesis. To improve the factual grounding of these hypotheses, we integrate the knowledge source, $\mathcal{K}$, using RAG \cite{lewis2020rag} for fact-based augmentation. To further refine the structure of $\mathcal{H}$, the \retriever clusters hypotheses based on their semantic and causal relationships, constructing a hierarchical tree structure $\mathcal{H}_t$. Within this tree, each node ${h}_t$ undergoes additional fact augmentation via RAG to ensure consistency and factual completeness. The entire process is formally represented as:

\begin{equation}
\mathcal{H}_t = \left\{\text{RAG}({h}_t, \mathcal{K}) \,\middle|\, {h}_t \in \text{T}\left(\bigcup_{i=1}^n \text{RAG}(\text{E}(d_i), \mathcal{K})\right)\right\},
\end{equation}

$\mathcal{H}_t$ denotes the final set of hypothesis tree. E and T denote the operations of extraction and tree structuring, respectively. $\text{RAG}({h}_t, \mathcal{K})$ denotes the operation of retrieval augmented generation, which will return top 5 related vectors from $\mathcal{K}$. The prompts of all operations can be found at Appendix~\ref{appendix_retriever'sprompt}

While it is desirable to utilize past materials from various client companies, 
not just those from the target client company, 
hypotheses from different industries and different challenges may not be particularly relevant 
for building hypotheses for the target client company. 
Therefore, we construct target hypotheses (hypotheses that apply to the target client company) 
from source hypotheses (hypotheses from other companies).
To this end, given a source hypothesis tree \(\mathcal{H}_t^s\) and a target 
hypothesis tree \(\mathcal{H}_t^t \), our goal is to iteratively transform \(\mathcal{H}_t^s\) 
into a valid \(\mathcal{H}_t^{t'}\). 
We use \(\mathcal{H}_t\) for tree-structured hypotheses, \(\mathcal{H}_t^s\) for source hypotheses,
\(\mathcal{H}_t^s\) for target hypotheses, and \(\mathcal{H}_t^{t'}\) for valid target hypotheses.
Using a database $\text{DB}$, queries are generated based on \(\mathcal{H}_t^s\) and executed to validate \(\mathcal{H}_t^t\). Using the success of the execution of the SQL code as a basis for judgment, this process is repeated until all queries are validated, and a final valid target hypothesis tree is obtained.
This process can be denoted as,
\begin{equation}
\mathcal{H}_t^{t'} = \bigcup_{h^t_t \in \mathcal{H}_t^t} \mathcal{U}\big(h^t_t, \text{EQ}(\text{GQ}(h_t^s, \text{DB}), \text{DB})\big)
\end{equation}

where \(\text{GQ}(h_t, \text{DB})\) generates a SQL query \(q\) for the hypothesis node \(h_t^s\), 
\(\text{EQ}(q, \text{DB})\) executes the query on the database \(\text{DB}\) and returns the result \(r \in \{\text{False}, \text{True}\}\), 
and \(\mathcal{U}(h_t^t, r)\) updates the hypothesis node \(h_t^t\) based on the query result \(r\).
$\mathcal{H}_t^{t'}$ is used as the few-shot prompt of \researcher.
The prompt can be found in Appendix~\ref{appendix_retriever'sprompt}

\section{Results}
\label{sec:experiments}

\subsection{Data and Models}
For all the experimental results reported here we have used a subset of Amazon's database of orders filled between the beginning of January and middle of November of 2024. The database was hosted on AWS S3 and was accessed via AWS Athena API. To protect privacy, companies and products have been anonymized throughout this paper. All the agents used the same LLM \\ \texttt{anthropic.claude-3-5-sonnet-20240620-v1:0} available through the AWS Bedrock service. Also, for the hyperparameters, please see Table \ref{label:HPs}.

\begin{table}[!ht]
\centering
\begin{tabular}{lr}
\toprule
Name & Value \\
\midrule
temperature (for report generation) & 0.8 \\
temperature (for reviewing and scoring) & 0.1 \\
maximum report length & 2500 \\
minimum number of SQL queries & 4  \\
maximum number of SQL queries & 8  \\
\bottomrule
\end{tabular}
\caption{Hyperparameters.}
\label{label:HPs}
\end{table}

%
%

\subsection{Experimental Settings}
In this paper we consider two business scenarios: "Sales optimization" and "Developing a product bundling strategy".

\textbf{Case 1: Sales optimization.} Here we consider a hypothetical scenario, where a major company (\emph{Company A}) that sells a diverse range of products through the Amazon marketplace, wants to maximize their market position and boost their sales in a particular product category. 
They would normally hire a professional consultant, who has access to relevant data sources, to conduct market research, including data collection, analysis and preparation of a comprehensive report with suggestions on how the client can achieve their objectives. However, consultants' work is usually time-consuming and might not be affordable to some. Here we show that \margen can automate the full cycle of consultants' work: from understanding the client's needs, to generating initial ideas, executing data queries, interpreting their results, and writing a human-readable report with actionable insights, supported by figures and data tables. 
We provide complete examples of such reports in Appendix \ref{appendix:sample_reports}.  


\textbf{Case 2: Developing a product bundling strategy.} In this scenario \margen is tasked to assist \emph{Company X}, an electronics company, to optimize their product bundling strategy, increase average order value and market share. \emph{Company X} also indicated that they are interested in understanding:

\begin{itemize}
    \item Which products are frequently purchased together
    \item How different customer segments respond to different bundle combinations
    \item What price points maximize bundle adoption while maintaining profitability
    \item How seasonal patterns affect bundle performance
    \item Whether certain bundles could help them capture market share from competitors
\end{itemize}

\section{Analysis}

\subsection{Does the \reviewer Really Helps the \writer to Improve the Reports?}
Yes. We noticed that the initial PDF reports usually have a lot of room for improvement. Specifically,

\begin{itemize}
    \item Sections tend to be too concise, leaving out important insights found by the \researcher;
    \item Some product/company names are not spelled in Latin characters;
    \item Recommendations are scant on specific implementation details;
    \item Technical terms are not spelled out (e.g what does the 'Personal Protective Equipment' product category include);
    \item Figures are very basic or not aesthetically pleasing (poor choice of fonts size, overlapping text, or axis labels, too big numbers. 
\end{itemize}

While most of the time the \writer is able to properly address the \reviewer's suggestions (especially when it comes to adding new or expanding existing textual content), we also observed failure cases. For example,

\begin{itemize}
    \item In response of the \reviewer's suggestion to enlarge a figure for better visibility, the \writer attempts to increase the resolution at which the figure is saved (using the \texttt{dpi} parameter in \texttt{plt.savefig} function). While this may improve the visibility, it doesn't directly address the \reviewer's suggestion. 
    \item The \reviewer sometimes wrongly suggests making changes to a figure in Section A, while in fact the figure referred to is in Section B (usually a neighboring section). 
\end{itemize}

\begin{figure}[h!]
   \centering
   \includegraphics[width=0.99\columnwidth] {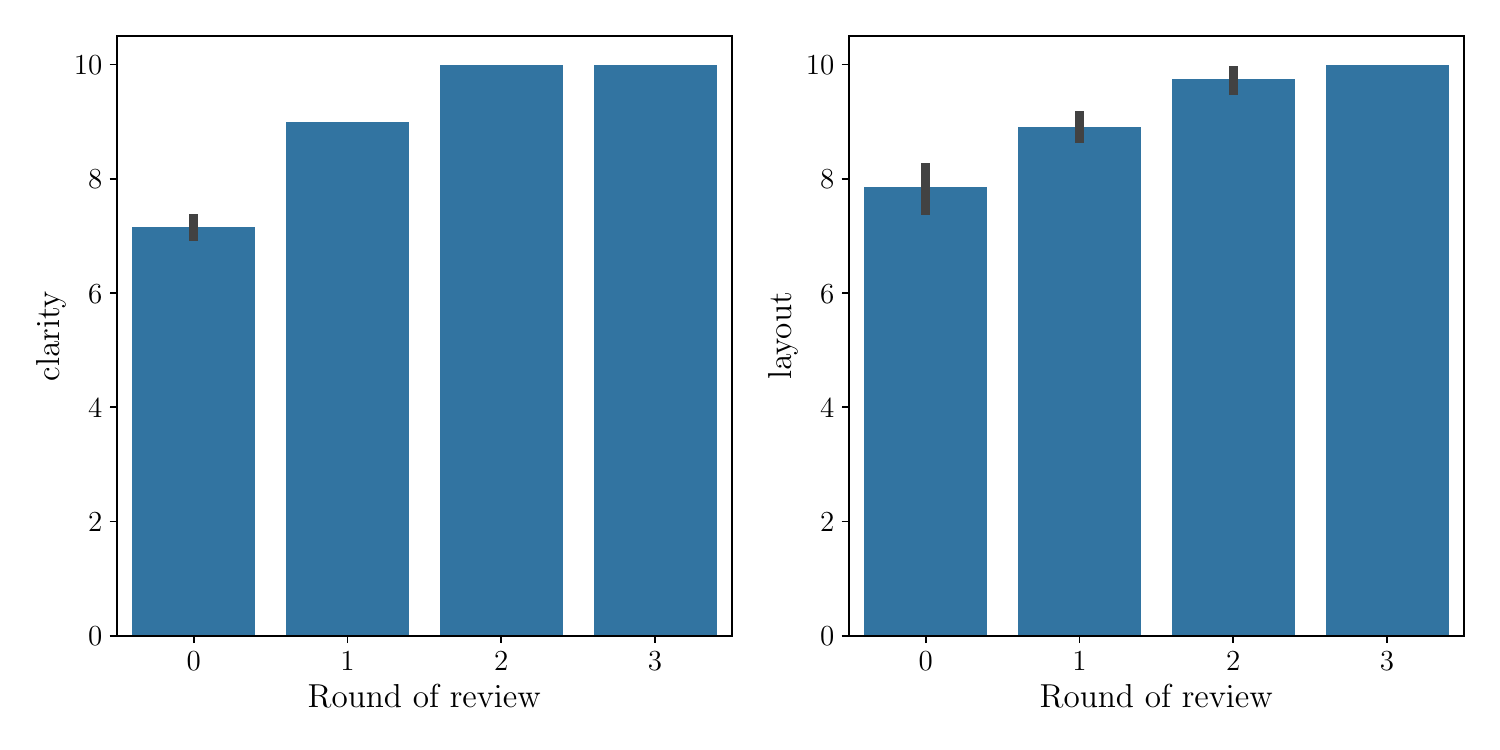}
   \caption{\label{fig:score_dynamics} Progressive improvement of scores over consecutive rounds as judged by the Reviewer. Bar height reflects an average over 20 reports. The error bars correspond to 95\% confidence intervals.}
\end{figure}

In general, a brief human inspection of the PDF reports from different rounds, and the correspondence between the \writer and \reviewer clearly shows that reports from later rounds rounds are clearly better than from earlier rounds. We note that it takes a maximum of 4 rounds (3 reviews) to reach the perfect score of 10 on both clarity and layout. Fig. \ref{fig:score_dynamics} suggests that initial reviews already have relatively high scores in the first round of review (around 7 for clarity and 8 for layout).

However, it is possible that the LLM simply increases the score because each new revision 
produced by the \writer is \emph{expected} to be better than the previous one, 
not because the new revision is objectively better. 
We have conducted additional experiment in order to rule out this possibility.
We run a set of pairwise comparisons between earlier and later versions of each report. 
Formally, we use an LLM that compares the same report in the $i$-th and $j$-th rounds and outputs $1$, if the $j$-th report is better than the $i$-th:

\begin{figure}[h!]
\centering
\includegraphics[width=0.99\columnwidth]{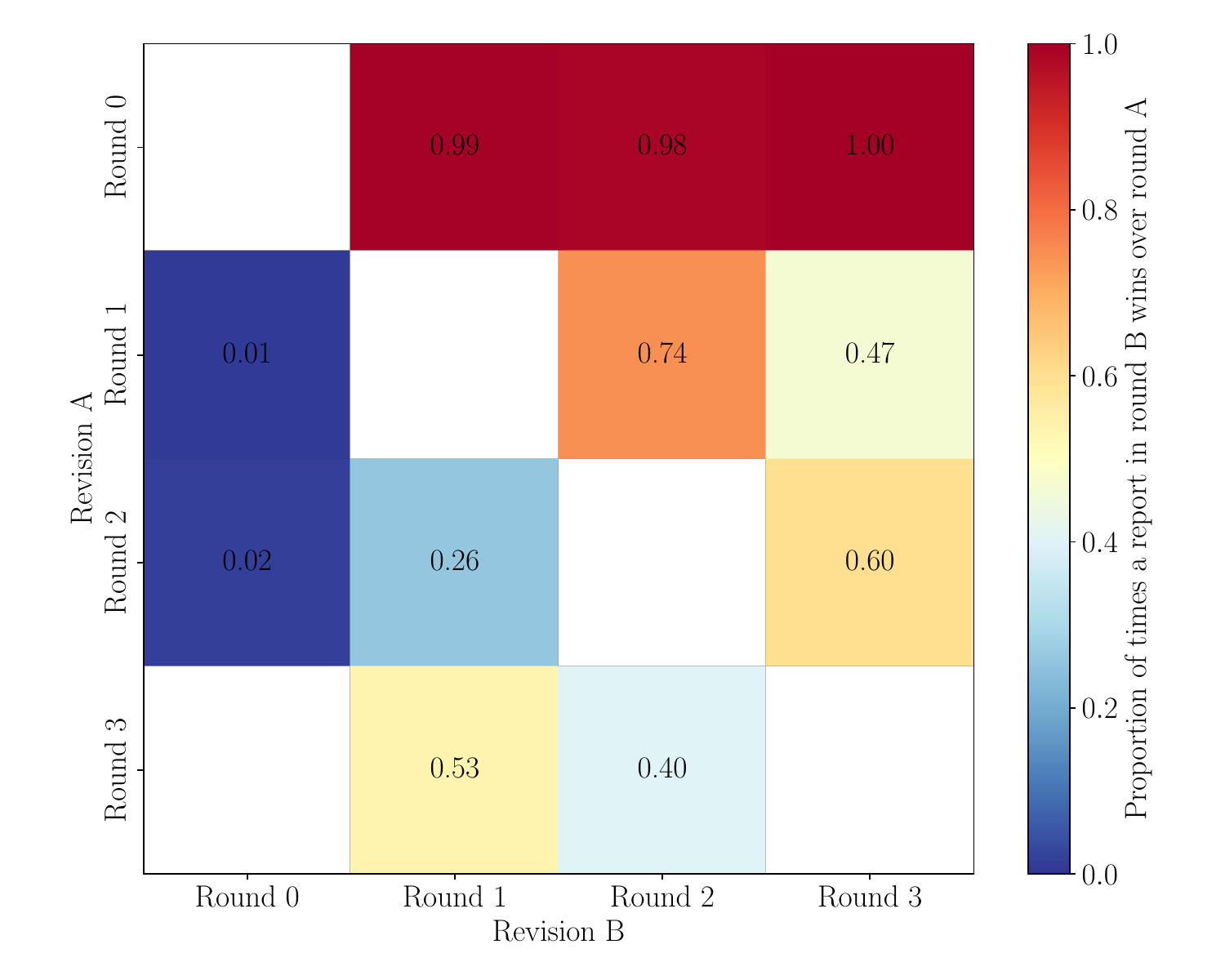}
\caption{\label{fig:all-vs-all} Pairwise comparison of the same report from different rounds of review. The numbers are averaged over 20 different reports.}
\end{figure}

\begin{equation}
f(R^{\text{pdf}}_i, R^{\text{pdf}}_j) = \begin{cases}
1 & \text{if } R^{\text{pdf}}_i \text{is worse than} R^{\text{pdf}}_j \\
-1 & \text{if } R^{\text{pdf}}_i \text{is better than} R^{\text{pdf}}_j \\
0 & \text{otherwise}
\end{cases}
\label{eq:pairwise}
\end{equation}

$\forall {i,j}$, s.t. $i < j$ and $i,j \in \{0,\dots,N\}$. Fig. \ref{fig:all-vs-all} shows that even when the LLM does not "know" which of the two reports correspond to which round, it still consistently judges \emph{later} revisions as better than those from earlier rounds.

\subsection{Can an LLM's Scores Be a Trusted Proxy for Human Quality Judgement?}

Our goal is to use an LLM for reviewing and scoring the generated reports and using those scores to guide their automatic iterative improvement. Before we can do that, we need to have confidence that the LLM's scores can be trusted as a proxy of human judgement about their quality. The most straightforward approach is \emph{LLM-as-a-judge}, in which each report is evaluated individually, that is \emph{not} relative to the quality of any other report. While \citet{lu2024aiscientistfullyautomated} claimed  that, at least when it comes to evaluating scientific papers, LLM-human agreement was higher than that between humans, \citet{si2024llmsgeneratenovelresearch} found that LLMs might still not be well-calibrated for this kind of evaluation of free-form text. For this reason, we also evaluate the reports' quality using a pairwise approach, similarly to \citet{si2024llmsgeneratenovelresearch}.

\textbf{Individual scoring}. In this type of scoring, each report ($N=20$) is submitted to the LLM or human scorer and evaluated \emph{independently}, i.e. not relative to any of the other reports. To assist in the scoring process, the scorers (both humans and LLMs) are instructed to provide a numerical answer on a scale of 1 to 10 (where 1 means "strongly disagree" and 10 means "strongly agree") to six statements. The statements are designed to evaluate the report's quality along a particular dimension: (1) non-triviality, (2) degree of justification, (3) clarity, (4) feasibility, 
(5) balance and (6) overall quality. 
See Appendix \ref{appendix:individual_questions} for the full prompt and Table \ref{table:scoring} for the scorers' instructions.

\begin{table*}[htbp]  
\centering
\caption{Scoring Criteria}
\label{table:scoring}
\begin{tabular}{|p{0.15\textwidth}|p{0.75\textwidth}|}
\hline
\textbf{Criterion} & \textbf{Description} \\
\hline
Non-triviality & The report goes beyond basic analysis, makes non-trivial conclusions/insights \\
\hline
Degree of justification & The conclusions are strongly supported by the data provided \\
\hline
Clarity & The report is well-organized, clear and easy to follow, always highlighting key insights \\
\hline
Feasibility & The recommendations in the report are specific, measurable and feasible \\
\hline
Balance & The report is well balanced, discusses multiple perspectives, and considers possible limitations and risks of following the suggested actions \\
\hline
Overall Score & This overall score is intended to reflect your overall quality judgement of a given report. The idea is that the 5 criteria listed above might not fully capture all the important aspects of quality (such as what weight should be given e.g. to clarity or another criterion). This "overall score" is not necessarily the sum/mean of the five other scores \\
\hline
\end{tabular}
\end{table*}

\textbf{Pairwise scoring.} Pairwise scoring is similar to Eq. \ref{eq:pairwise}, except that instead of comparing different revisions of the same report, we compare \emph{final} versions of different reports (i.e. generated from slightly different research histories, but same prompts). Specifically, given a set of $N$ reports $\{R_i^\text{pdf}\}_{i=1}^N$, we prompt the LLM to decide which of the two reports, $R^{\text{pdf}}_i$ and $R^{\text{pdf}}_j$ is better. To reduce variance, each pair gets to be compared $K$ times. The results of this comparison are recorded to the matrix $M^{N \times N \times K}$, where the entry $M_{ijk}$ represents the result of the comparison defined as follows:

\begin{equation}
M_{ij} = \begin{cases} 
1 & \text{if } R^{\text{pdf}}_i \text{ is better than } R^{\text{pdf}}_j, \\ 
-1 & \text{if } R^{\text{pdf}}_j \text{ is better than } R^{\text{pdf}}_i, \\ 
0 & \text{if there is a tie}, \\ 
0 & \text{if } i = j \text{ (no comparison with itself)}. 
\end{cases}
\end{equation}

\begin{figure}[h!]
\centering
\includegraphics[width=0.99\columnwidth]{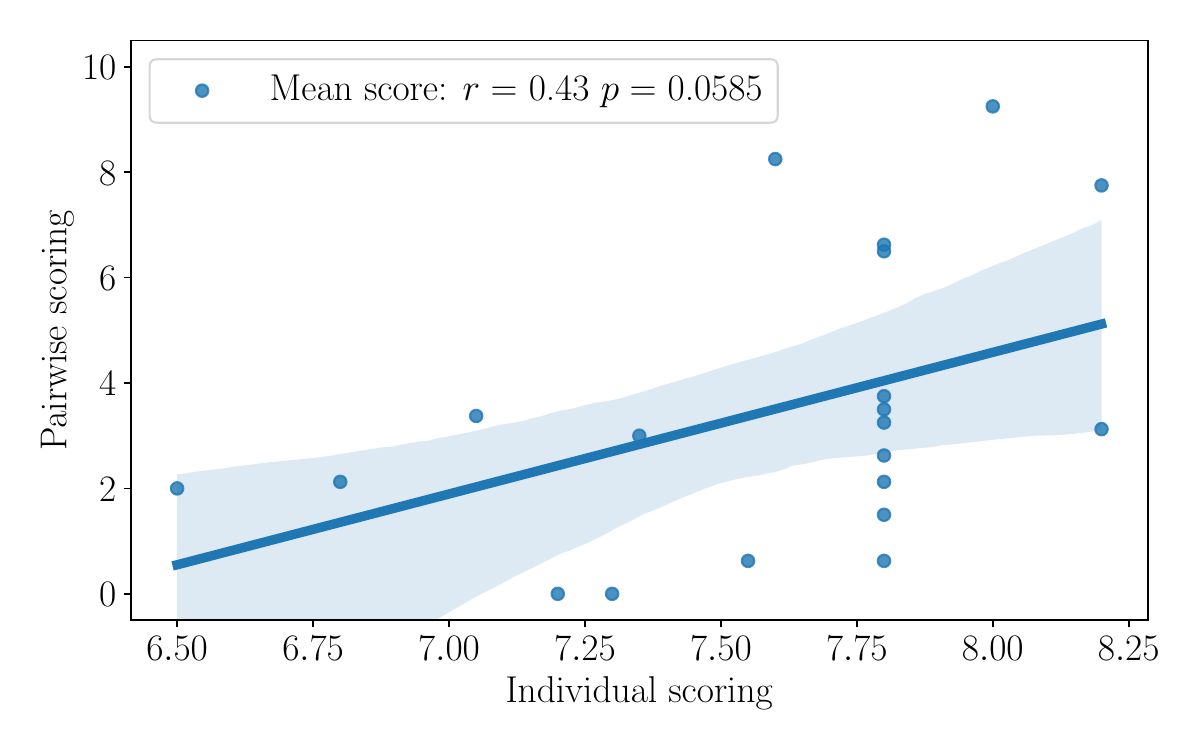}
\caption{\label{fig:pairwise-final} Final report quality scores obtained using LLM-based individual and pairwise scoring are somewhat correlated, but individually assigned scores range between 6.5 and 8.25, potentially failing to capture part of the quality variability.}
\end{figure}

The scores of each report are obtained by averaging $M$ along the third and then second dimensions. Intuitively, these scores represent the proportion of times in which the $i$-th report won over the other. Fig. \ref{fig:pairwise-final} shows that both approaches to scoring reviews mostly agree (with Pearson $r = 0.43$, $p = 0.0585$). The fact that AI scores for the same report obtained using two different automatic reviewing approaches are correlated, suggests that the LLM is relatively consistent in its perception of quality regardless of the method used. However, when using individual evaluation, the scores tend towards the middle of the scale (notice that individual scores are concentrated between 6.5 and 8.25), potentially failing to capture part of the reports' quality variability (Fig. \ref{fig:pairwise-final}). For this reason, unless specified otherwise, we report result obtained using this pairwise evaluation method.

\textbf{Human scoring}. To assess the agreement between the LLM's judgement of report quality ("AI scores") and that of an average human, we asked two human reviewers to score the same $20$ reports as those scored by the LLM. Each of the 20 reports was scored once by each reviewer, and the scores were averaged between reviewers.

\begin{figure}[h!]
\centering
\includegraphics[width=0.99\columnwidth]{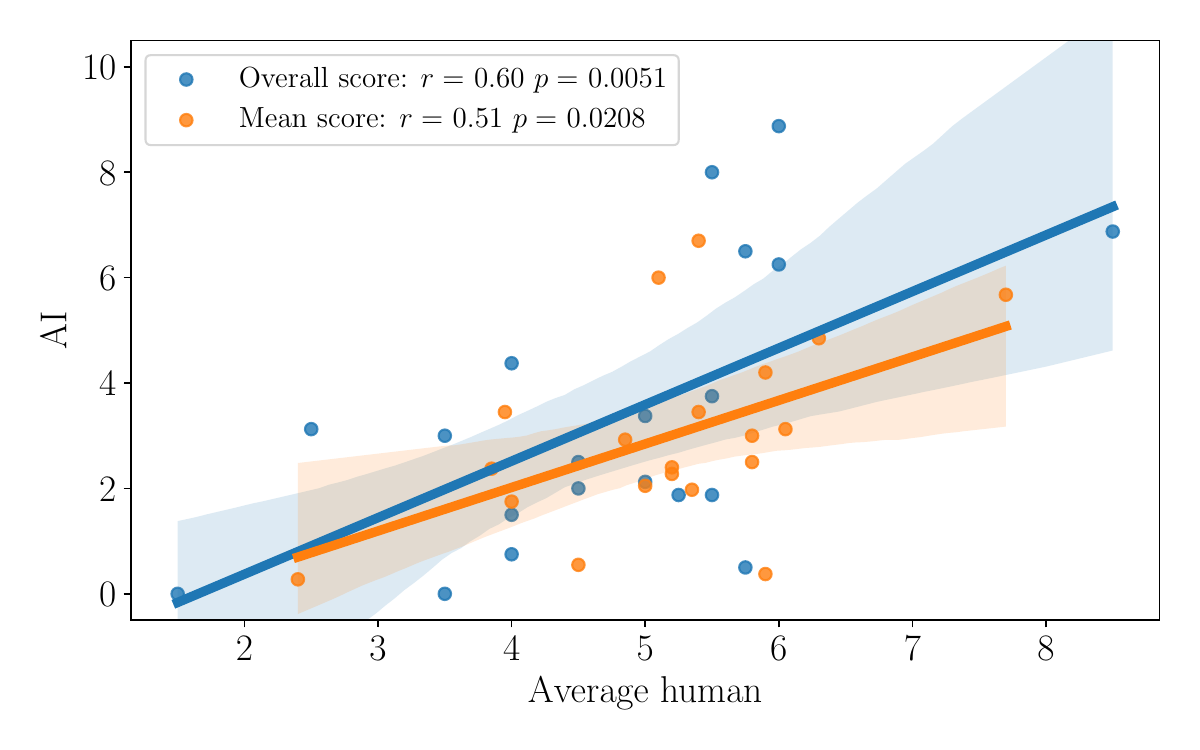}
\caption{\label{fig:avg_hum_vs_pwAI} The correlation between average human and AI-generated (pairwise) quality scores for 20 final reports.}
\end{figure}

Fig. \ref{fig:avg_hum_vs_pwAI} shows that the average human scores for overall quality correlate well with pairwise AI scores (but see also correlations for individual AI scores, as well as detailed criterion-wise correlations in Appendix \ref{appendix:supplem_figs}). This is important, because we can further improve the quality by generating $N$ reports in parallel with the same prompt, and while the reports will be mostly similar, given a reliable automatic reviewer, we can select the best of them as the final one.


\subsection{Does the Injection of Expert Knowledge by \retriever Improve the Report Quality?}
\begin{figure}[t!]
\centering
\includegraphics[width=0.99\columnwidth]{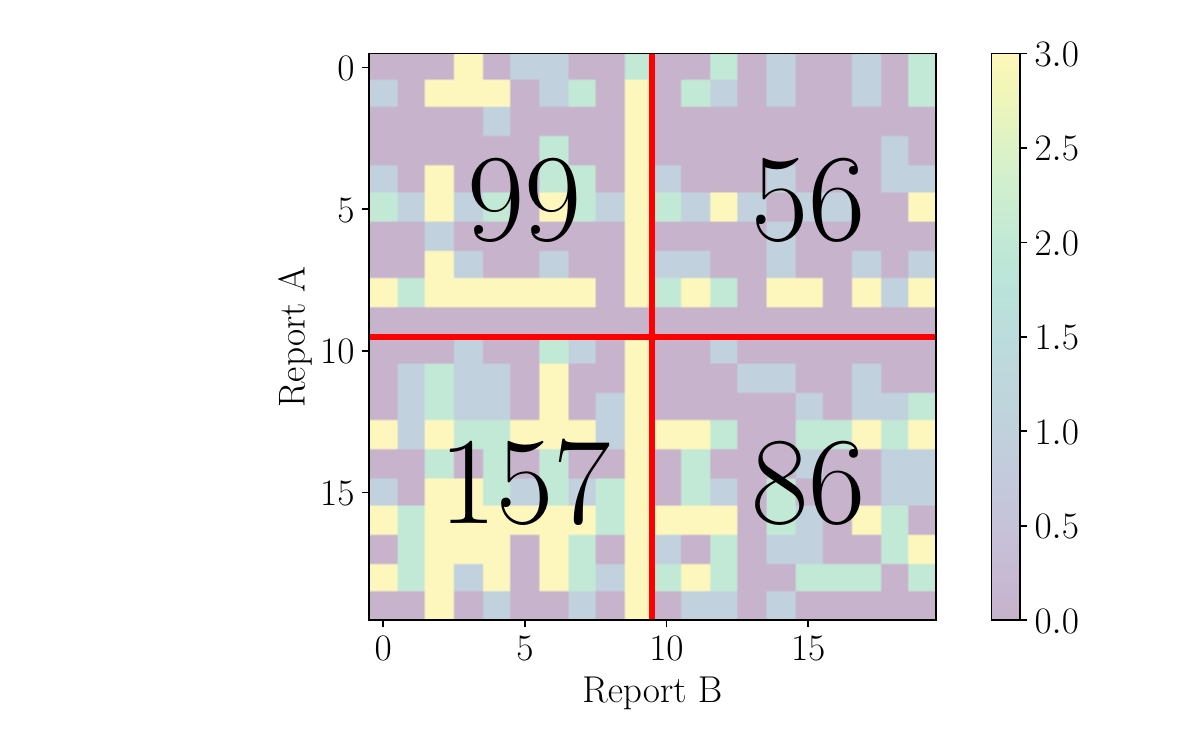}
\caption{\label{fig:tree-vs-notree} Pairwise comparison of \margen reports with and without added examples of expert reasoning. Each pixel shows the number of times report A was judged as better than report B in a pairwise comparison. Reports 0 - 9 and 10-19 were generated without and with the domain knowledege provided by the \retriever, respectively. Reports 10 - 19 (generated with the added few-shot examples generated by the \retriever) are clearly better than reports 0 - 10 (generated without that few-shot examples).}
\end{figure}

As mentioned in Section \ref{subsec:Retriever}. In many real-world scenarios, a large number of unstructured expert documents already exist. 
We aim to fully exploit this wealth of information to enhance the quality and comprehensiveness of the generated reports. 
We randomly selected five companies, 
built their own \retriever, and generated 20 reports for each: 10 with added few-shot examples generated by the \retriever and 10 without them. 
Fig. \ref{fig:tree-vs-notree} shows that the proposed \retriever significantly enhances the quality of generated reports and demonstrates strong potential for application in real-world business scenarios. 


\section{Limitation and Future Works}
We mention several limitations and future work.
First, while we have shown above that LLMs can be a trusted proxy for human perception 
of quality, at least on the specific dimensions of quality that we defined for our analysis, 
and that LLMs can improve the quality of reports 
by themselves, we have not conducted evaluations on whether the market reports 
created by \margen are equivalent to or better than those created by humans. 
This point is important future work.
Also, regarding whether the data extraction and analysis performed by \margen is 
correctly implemented, we would like to mention that users can verify this because all 
executed SQL queries and other operations can be checked.
We have also conducted preliminary analysis on validating the data provided by \margen using 
another agent called \textsc{Verifier} (details can be found in the Appendix~\ref{appendix:verifiers_prompt}).
Incorporating a data validation process could lead to building a more reliable system.

\section{Conclusions}
\label{sec:conclusions}
In this paper, we introduced \margen, a framework for automated market research that leverages 
LLMs to generate data-driven insights. 
Our system demonstrates that \margen can effectively conduct end-to-end market research, 
from initial data exploration to producing comprehensive market reports with visualizations.
Experimental result showed alignment between expert human and AI quality assessments, 
suggesting that automated evaluation can serve as a reliable proxy for human judgment, enabling efficient quality control at scale.
Additionally, we showed that \margen can improve the quality of reports through iterative review and revision processes. 
Furthermore, we demonstrated that report quality can be improved by utilizing professional consultants' unstructured knowledge documented in PowerPoint through in-context learning.
While future work remains, we believe our work represents an important step toward democratizing access to professional-grade market research. 

\bibliographystyle{ACM-Reference-Format}
\bibliography{bibliography}

\onecolumn
\newpage

\twocolumn
\appendix
\section*{Appendix}


\section{Supplementary Figures}
\label{appendix:supplem_figs}

\begin{figure}[ht!]
\centering
\includegraphics[width=0.8\columnwidth]{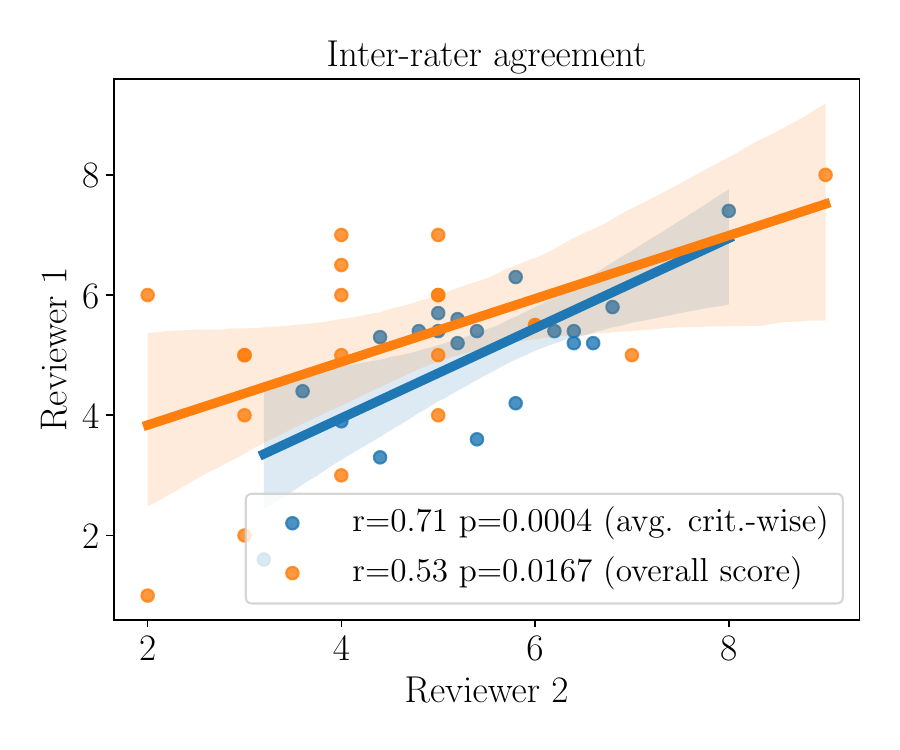}
\caption{\label{fig:supplem0} Inter-rater agreement. The orange line and data points correspond to overall scores (which are not necessarily the average of the other five criteria (see Appendix \ref{appendix:scorers_instructions}). The blue line and data points correspond to the simple averages of the 5 criteria (NT, DJ, CL, FE and BA).}
\end{figure}

\begin{figure*}[ht!]
\centering
\includegraphics[width=0.9\textwidth]{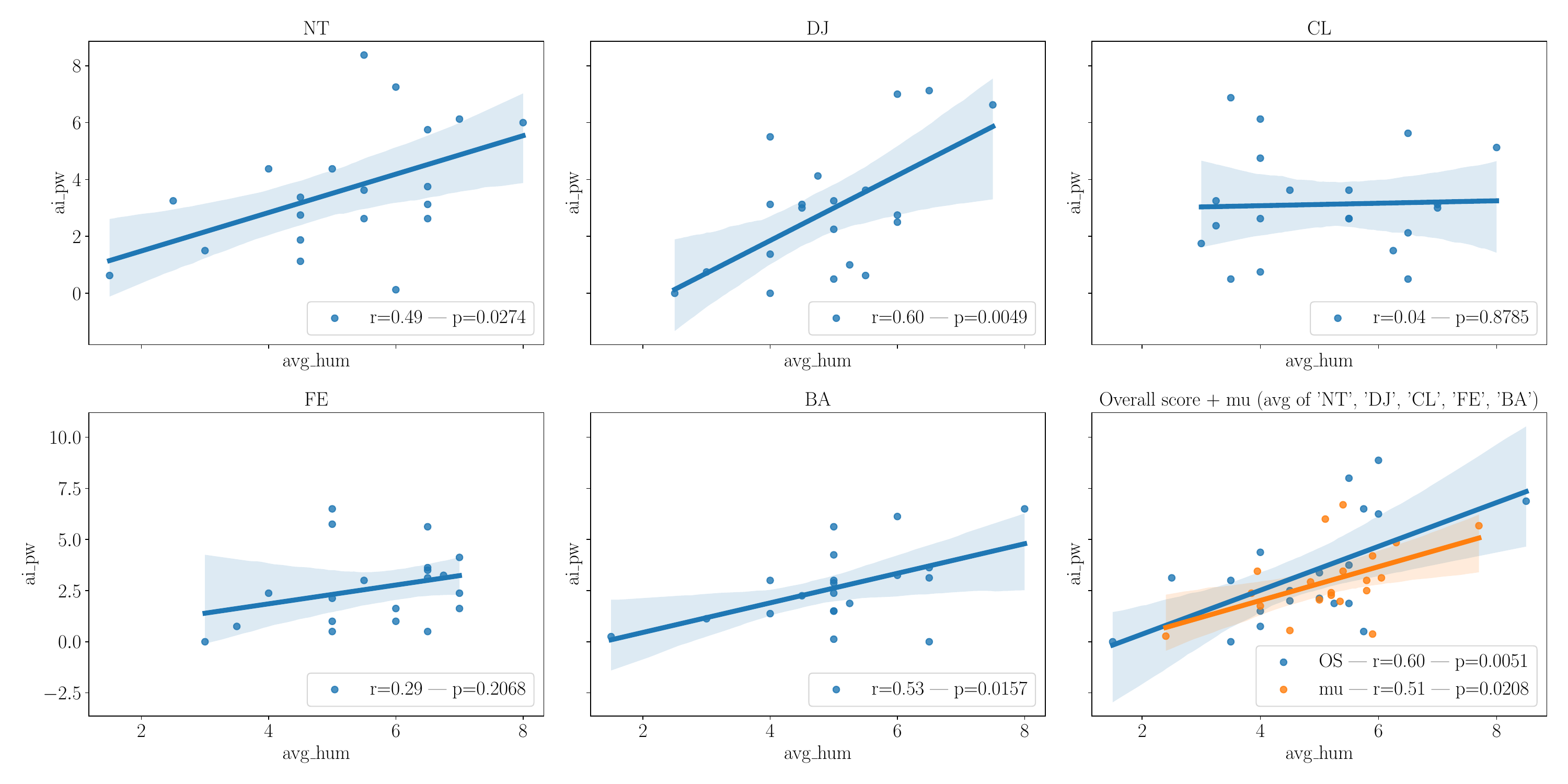}
\caption{\label{fig:supplem1} Analysis of the agreement between human and AI-generated (pairwise) quality scores. Each panel shows the correlation for one of the criteria. The bottom right panel additionally shows the correlation between human and the mean of the AI-generated scores on the following five criteria: NT, DJ, CL, FE and BA.}
\end{figure*}

\vfill

\begin{figure*}[ht!]
\centering
\includegraphics[width=0.9\textwidth]{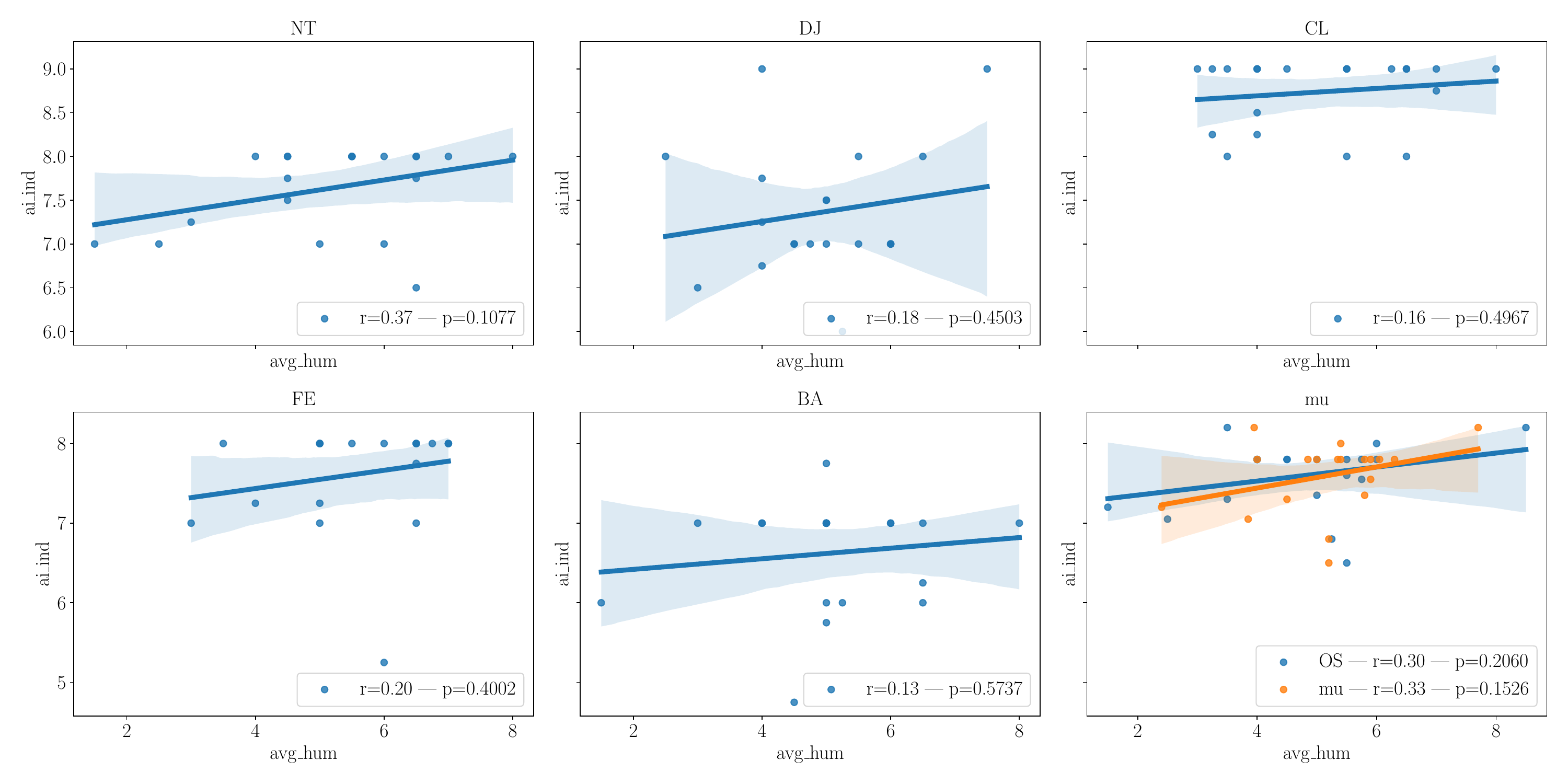}
\caption{\label{fig:supplem2} Analysis of the agreement between human and AI-generated (individual) quality scores. Each panel shows the correlation for one of the criteria. The bottom right panel additionally shows the correlation between human and the mean of the AI-generated scores on the following five criteria: NT, DJ, CL, FE and BA.}
\end{figure*}
\clearpage

\section{Prompts}
Product names and other information have been anonymized as appropriate.
\label{appendix:prompts}

\subsection{Researcher's Prompt Template}
\label{appendix:prompts:researcher}

The template below accepts 3 string parameters: \texttt{CLIENT\_COMPANY}, \texttt{CLIENT\_PROVIDED\_INFO}, \texttt{DB\_SCHEMA}, \texttt{MAX\_QUERIES}, and \texttt{MAX\_QUERIES}.

For Case Study 1 they were as follows:

\begin{lstlisting}

CLIENT_COMPANY = "Client 123456789"
CLIENT_PROVIDED_INFO = "
- the client is looking to optimize theirs sales;
- the client has suggested that they want to become the leader in the category 'category A'."

MAX_QUERIES = 4
MAX_QUERIES = 8



\end{lstlisting}

\begin{lstlisting}

# Situation

Your client is {CLIENT_COMPANY}. 
In your initial discussion with the client, you have established the following:
{CLIENT_PROVIDED_INFO}

# Task

You task is to conduct research, within the constraints listed, to help your customer achieve their goals. Specifically you must 
1) come up with an initial hypothesis for 
    - what actions you client can take to optimally achieve their goals;
    - what might be the possible risks or tradeoffs associated with following your suggestions
2) obtain relevant data and see if they support your hypothesis; if the data does not support it, adjust or change your hypothesis and validate it again against data.

# Data Sources

You have access to a database table, which you can query it using standard SQL syntax. The database table has the following schema:

{DB_SCHEMA}

# Constraints

1) When (and if) you need to get relevant data, write your SQL query surrounded with triple backticks, for example:

```
{EXAMPLE_SQL}
```

2) For all your queries, only request data from 2024.
3) Do not write SQL queries that are likely to return more than 200 rows, and thus use aggregation operations as appropriate.
4) Never generate more than one SQL query at a time and never proceed to furter reasoning or analysis before the result of the query is returned to you.
5) If you are given any names (company, product etc.), always use them exactly, or your queries will return no results.
6) You must complete your research using no more than {MAX_QUERIES} (but no fewer than {MIN_QUERIES}) queries.
7) To select rows from 2024, always use the more efficient `WHERE EXTRACT(YEAR FROM COLUMN_X) = 2024`, not `WHERE DATE_FORMAT(COLUMN_X, '%Y') = '2024'`.
8) The actions in your final report must not require any further research or data analysis.


# Additional Instructions

Each time you make a query, I will return its result to you in JSON format for your analysis and consideration.
Except for the constraints given, you have unlimited creative freedom to tackle this task.
If the retrieved data does not support the initial hypothesis, revise it and continue your research.


# Expected Result

When your proposed hypothesis is strong enough, write an exceptionally well-written professional report of no longer than {REPORT_MAX_WRDS} words. 
This report must be surrounded by the <FINAL_ANSWER> and </FINAL_ANSWER> tags, for example:

<FINAL_ANSWER>
Your final answer goes here.
</FINAL_ANSWER>

The report MUST cover the following points:

1. Executive Summary
   - Provide a concise overview of key findings and recommendations
   - Highlight 3-5 main points that capture the essence of the report

2. Introduction
   - State the purpose of the research
   - Outline the scope and objectives
   - Briefly describe the methodology used

3. Market Overview
   - Analyze the current state of the market
   - Identify key players and their market shares
   - Discuss market size, growth rates, and trends

4. Competitive Analysis
   - Identify main competitors
   - Analyze their strengths and weaknesses
   - Compare the client's position relative to competitors

5. Customer Analysis
   - Define target customer segments
   - Analyze customer behavior, preferences, and needs

6. Product Analysis
   - Evaluate current product offerings
   - Analyze performance of different product lines or segments
   - Identify areas for improvement or expansion

7. Pricing Analysis
   - Analyze the customer's current pricing strategies in comparison with those of competitors
   - Identify optimal price points for different customer segments

9. SWOT Analysis (note: include only when you have the data for this)
   - Strengths: Internal positive attributes
   - Weaknesses: Internal areas for improvement
   - Opportunities: External factors that could benefit the business
   - Threats: External factors that could harm the business

10. Recommendations
    - Provide actionable strategies based on findings
    - Include short-term and long-term suggestions

11. Implementation Plan
    - Outline steps to implement recommendations
    - Provide timeline and milestones
    - Identify potential challenges and mitigation strategies

12. Financial Projections
    - Estimate potential impact of recommendations
    - Provide ROI analysis where applicable

13. Conclusion
    - Summarize the main points
    - Highlight critical issues and opportunities
    - Reinforce the value of implementing recommendations


And adhere to the following Best Practices:

1. Use a mix of quantitative and qualitative research methods
   - Analyze sales/orders data, market trends, and customer behavior

2. Segment analysis where appropriate
   - Break down findings by product categories, customer segments, or other relevant factors

3. Use clear, concise language
   - Avoid jargon unless necessary
   - Define technical terms when used

4. Be specific and actionable in recommendations
   - Provide clear next steps
   - Explain expected outcomes of implementing recommendations

5. Use a logical flow
   - Ensure each section builds on previous ones
   - Use transitions to connect different parts of the report

6. Incorporate client's perspective
   - Address specific concerns or questions raised by the client
   - Align recommendations with client's goals and capabilities
\end{lstlisting}

\subsection{Writer's Prompt}

\subsubsection{Writer's Initial Prompt Template}
This prompt template is used to build the initial prompt for the Writer, i.e. when no review has been received yet.

\begin{lstlisting}
{history}

What you see above is the full history of a chat between and an AI assistant and a human. The AI assistant has conducted market research and analysed the results. Based on this history, write a concise, but professional, consulting report in markdown format.
In this new report, each finding must be illustrated by a corresponding figure (one figure per SQL query response), inserted in its appropriate place. The png files for figures don't exist yet, so instead of adding a link to the image file in the markdown figure, you need to insert python code blocks, one for each figure. Each code block must appear in its appropriate place in the report and be sufficient to generate the corresponding figure and save it in png format. Each code block must be followed by corresponding figure caption concisely explaining what the (future) figure will show. Do not add titles to the figures. Do not use `ha` as a keyword argument for `tick_params`.  Avoid using pandas. Each python code block must start with the following lines:

import matplotlib.pyplot as plt
import numpy as np
plt.rcParams["font.family"] = "Arial Unicode MS"  # (!!!) needed to properly render Japanese characters


and end with the following two lines:

plt.tight_layout()
plt.savefig('fig_ID.png')

where ID is the ordinal number (integer) of the figure in the report.
\end{lstlisting}

\subsubsection{Writer's Later Prompt Template}
This prompt template is used to build the later prompt for the Writer.

\begin{lstlisting}
Thank you. A professional editor at Nature Publishing Group has reviewed the document you created.
Here's their review:

<REVIEW_ROUND_{round}>
{review}
</REVIEW_ROUND_{round}>

Please address these reviewer's concerns and comments by editing your previous markdown and python code in it.
First generate a new markdown using the previous instructions. In the future, you will be prompted to generate the corresponding new latex version.
\end{lstlisting}

\subsubsection{Writer's Response to Reviewer Template}
After generating an improved version of the report (in markdown), which addresses the Reviewer's comments and suggestions, the Writer is prompted to write a brief response to the Reviewer:

\begin{lstlisting}
Now write a short response to the reviewer describing the specific things you did to address their concerns and comments.
The part of your answer that will be sent back to the reviewer must be surrounded by <RESPONSE_TO_REVIEWER> and </RESPONSE_TO_REVIEWER>, for example:

<RESPONSE_TO_REVIEWER>
List the things you did.
</RESPONSE_TO_REVIEWER>  
\end{lstlisting}

\subsubsection{Writer's Figure Template}
This prompt template is used to build figure.

\begin{lstlisting}
\begin{figure}[h!]
\centering
\includegraphics[width=0.8\textwidth] {fig_1.png}
\caption{\label{fig:1} Caption for Figure 1 from the markdown document.}
\end{figure}
\end{lstlisting}

\subsubsection{\writer's Latex Generation Template}
After generating the report in markdown, the Writer is prompted to generate a latex version of it:

\begin{lstlisting}
Now convert this report into latex code using the latex template (see below), but keep the structure and content exactly the same as in the markdown document. Importantly, replace the python blocks with latex code for inserting the corresponding figure. For example for Figure 1, the latex code should be as follows:

{figure_example}

Here's the latex template:

{latex_template}

In your answer, the latex code of the entire report must be written in one block and surrounded with triple backticks (```). Also, be sure to escape the ampersand (`&`) with a backslash or the latex will not render (for example, instead of `R&D` and similar, write `R\&D`).
\end{lstlisting}

\subsection{\reviewer's Prompts}
\label{appendix:reviewers_protmps}

\subsubsection{\reviewer's System Message}
This is a system message.

\begin{lstlisting}
You are a highly experienced marketing consultant with decades or professional experience. You teach consulting at the Harvard Business School.
\end{lstlisting}

\subsubsection{Reviewer's Initial Prompt Template}
This prompt template is used only for the first round.

\begin{lstlisting}
Your task is to review and score the attached report thoroughly, paying special attention to whether the findings are presented clearly and the visuals are easy to understand and support the narrative. Write your brief feedback (text of the review) covering specific improvement suggestions. The scores must reflect the quality of the report on "clarity" and "layout" (on a 10-point integer scale, where 1 is bad and 10 is perfect), be surrounded by triple backticks, and sctrictly follow the format below:

```json
{{
    "clarity": INT, 
    "layout": INT
}}
```

Here are some questions to help you guide (but not limit or otherwise restrict) your review:

Did the authors follow the best practices, specifically:

1. Did they use a mix of quantitative and qualitative research methods?
   - Did they analyze sales/orders data, market trends, and customer behavior?

2. Did they segment analysis where appropriate?
   - Did they break down findings by product categories, customer segments, or other relevant factors?

3. Did they use clear, concise language
   - avoiding jargon unless necessary?
   - defining technical terms when used?

4. Were they specific and actionable in their recommendations?
   - Did they provide clear next steps?
   - Did they explain expected outcomes of implementing recommendations?

5. Did they use a logical flow
   - ensuring each section builds on previous ones?
   - using transitions to connect different parts of the report?

6. Did they incorporate client's perspective?
   - addressing specific concerns or questions raised by the client?
   - aligning recommendations with client's goals and capabilities?
\end{lstlisting}

\subsubsection{\reviewer's Later Prompt Template}

This template is used for all versions except the first one. Note that only the \writer's response to the \reviewer is pasted into the template, while the updated report is added to the prompt as a sequence of images.

\begin{lstlisting}
The creator of the report has addressed your comments and suggestions and sent you its updated version. They also added some comments to highlight what they changed:

<CREATOR_RESPONSE>
{response}
</CREATOR_RESPONSE>

Please review the report again, paying special attention to whether your suggestions and comments have indeed been addressed satisfactorily. If they have been addressed insufficiently, point this out in your review. If your previous suggestions have been addressed partially, consider raising your score proportinately to the degree and quality of improvements. Your review and scores must follow the same format as before.

\end{lstlisting}

\subsubsection{Prompt For Scoring Reports Individually}
\label{appendix:individual_questions}
This prompt is used to score reports individually.

\begin{lstlisting}[xleftmargin=0pt]
You are given images of the final PDF report written by a consultant at Amazon.
Before writing their report, the consultant received the following instructions:

<CONSULTANTS_INSTRUCTIONS>
{creators_prompt}
</CONSULTANTS_INSTRUCTIONS>

Following these instructions, the consultant did some research, the full history of which is below:

<CONSULTANTS_REPORT>
{consultants_report}
</CONSULTANTS_REPORT>

Read the report carefully and fill out the following questionnaire in JSON format as illustrated below.

```json
{{
"The report goes beyond basic analysis, makes non-trivial conclusions/insights.": INT,
"The conclusions are strongly supported by the data provided.": INT,
"The report is well-organized, clear and easy to follow, always highlighting key insights.": INT,
"The recommendations in the report are specific, measurable and feasible.": INT,
"The report is well balanced, discusses multiple perspectives, and considers possible limitations and risks of following the suggested actions.": INT
}}

As you do so, approach scoring the report with an unbiased perspective, while holding it to an exceptionally high standard of quality. Your answers must be on a 10-point integer scale from 1 (strongly disagree) to 10 (strongly agree).

\end{lstlisting}

\subsubsection{Prompt For Pairwise Scoring of Reports}
\label{appendix:pairwise_questions}
This prompt is used to evaluate reports pairwise.

\begin{lstlisting}

You are given two reports independently written by two different consultants (one consultant - one report). These consultants had exactly the same background information, instructions and access to the same data. Before writing their reports, the consultants received the following instructions:

<CONSULTANTS_INSTRUCTIONS>
{creators_prompt}
</CONSULTANTS_INSTRUCTIONS>

The two reports for you to judge are as follows:

<REPORT_A>
{report_A}
</REPORT_A>

<REPORT_B>
{report_B}
</REPORT_B>

Using the specific instructions and general guidelines below, judge which report is better (i.e is the winner). If both of them are equally good (or bad), the outcome of our judgement is a "DRAW".

**Specific Instructions**:

1. Check for alignment with the original problem description:
   - Does the report directly address the client's objectives and needs?
   - Are all key questions answered?

2. Assess the methodology:
   - Are the research methods used appropriate?
   - Are the sample sizes sufficient? 

3. Examine data analysis:
   - Does the report contain appropriate statistical analyses?
   - Are the conclusions strongly supported by the data?
   - Are there any potential biases in the analysis?
   - Are the conclusions sufficiently non-trivial?
   - Is the report Are the conclusions sufficiently non-trivial?

4. Review the presentation of findings:
   - Does the report present information clearly and in a logical order?
   - Are the key insights clearly highlighted and easy to understand?
   - Does the report discuss the limitations in the analysis?

5. Evaluate the actionability of recommendations:
   - Are the recommendations specific, measurable, and feasible?
   - Are the connections between findings and recommendations clear?

As you write up your review, follow the general guide lines below:

**General Guidelines**:

1. Maintain objectivity:
   - Approach each report with an unbiased perspective
   - Focus on the quality of work rather than what might be perceived as personal preferences

2. Consider the target audience:
   - Evaluate if the report is tailored to the client's level of expertise (very basic understanding of marketing, data science)
   - Assess the overall readability and accessibility of the content

3. Look for innovation and creativity:
   - Identify unique approaches or insights that set reports apart
   - Recognize consultants who go beyond basic analysis

4. Assess the overall professionalism:
   - Evaluate the overall polish and presentation of the report

5. Provide constructive feedback:
   - Highlight strengths as well as weaknesses in each report

6. Consider the client's perspective:
   - Assess how well each report addresses the client's specific needs
   - Evaluate the potential impact of findings on the client's business


Start by writing a review assessing each of the reports using the instructions and guidelines above. Finish your answer with your final judgement, which must be in the following format:

```json
{{"winner": "REPORT_A" OR "REPORT_B" OR "DRAW"}}
```
\end{lstlisting}

\subsection{Verifier's Prompt}
\label{appendix:verifiers_prompt}
This template is used to verify data analysis.

\begin{lstlisting}
# Situation

Another consultant has conducted market research and written a report (see images or each page and hi-resolution images for each figure in it above). The client, for whom the report was written, has voiced concerns that some data presented in the report may not be entirely accurate.

For your reference, you are also given the latex version of the report:

<REPORT_LATEX>
{latex}
</REPORT_LATEX>

# Task

You task is to double-check all the numberical data contained in the given report.

# Data Sources

Just like the other consultant, you have access to a database table, which you can query using standard SQL syntax (Presto SQL used in AWS Athena). The database table has the following schema:

{DB_SCHEMA}

For your reference, here's the full history of the other consultant's research, where you can see what SQL queries he made. If you don't see anything wrong with thouse queries, you can use the same or similar SQL queries to double-check the results that the consultant's report contains.

<FULL_RESEARCH_HISTORY>
{final}
</FULL_RESEARCH_HISTORY>

# Constraints

1) When (and if) you need to get relevant data, write your SQL query surrounded with triple backticks, for example:

```
{EXAMPLE_SQL}
```

2) For all your queries, only request data from 2024.
3) Do not write SQL queries that are likely to return more than 200 rows, and thus use aggregation operations as appropriate.
4) Never generate more than one SQL query at a time and never proceed to furter reasoning or analysis before the result of the query is returned to you.
5) If there are any names (company, product etc.), always use them exactly, or your queries will return no results.
6) To select rows from 2024, always use the more efficient `WHERE EXTRACT(YEAR FROM COLUMN_X) = 2024`, not `WHERE DATE_FORMAT(COLUMN_X, '%Y') = '2024'`.
7) Do not offer any analysis, judgement or review of the quality or the report. Only look for inaccuracies and report them (if any).
8) When your queries contain calculations, cast all numbers to DOUBLE, and round the result to 4 decimal places, for example:
  `ROUND(CAST(SUM(COLUMN_X) AS DOUBLE) / NULLIF(CAST(SUM(SUM(COLUMN_X)) OVER () AS DOUBLE), 0) * 100.0 AS market_share_percentage, 4)`


# Additional Instructions

Each time you make a query, I will return its result to you in JSON format for your analysis and consideration.
Each time you get the data from me, if there is a discrepancy between the number(s) I return and the number(s) you see in the other consultant's report, write:

The [revious consultant said the DATA_POINT_NAME was CONSULTANT_VALUE, but according to your data it is actually ACTUAL_VALUE. (replace DATA_POINT_NAME, CONSULTANT_VALUE and ACTUAL_VALUE with the corresponding name and values).


For example: 

"Previous consultant the total market size was 3.1T yen, but according to your data it is actually 2.1B yen."


Except for the constraints given, you have unlimited creative freedom to tackle this task.


# Expected Result

When you have checked all the numberical data, concisely list the specific errors and/or inaccuracies you have found (if any), also mentioning their known (or likely) causes. At the very end, summarize the incorrectly reported data in a markdown table as follows:

| data        | reported | actual |
|-------------|----------|--------|
|sales        | 3000     |4000    |
| ...         | ...      | ...    |
|Client share | 30%      |34%     |
\end{lstlisting}

\subsection{Retriever's Prompt}
\label{appendix_retriever'sprompt}
This template is used for the Retriever.

\subsubsection{Extraction.1}
This template is used to extract relevant questions and hypotheses from images.
\begin{lstlisting}  
  Now you're a consultant who consults for retail companies. Analyze the provided images and extract relevant questions and hypotheses.

Instructions:
0. Even if the content is Japanese, try to understand and analyze it.
1. For each image, identify the key questions and hypotheses that can be extracted from the content. 
2. If an image does not contain any useful information for companies, simply return "None" for that image.
3. Structure your output as follows:

[{
    "question": ,
    "hypothesis": 
}]

every Questions follows one Hypotheses. The output should be provided in English.
**IMPORTANT**
Just output the json format result directly. DO NOT add additional explanations or introduction in the answer unless you are asked to. Make sure the only keys in the JSON object are "question", "hypothesis".

\end{lstlisting}

\subsubsection{Extraction.2}
This template is used to extract key questions and hypotheses.

\begin{lstlisting}
You are an experienced retail consultant specializing in data analysis to extract key questions and hypotheses.

Please carefully review the provided data and perform the following tasks:

1. Filter irrelevant content: Remove any question and hypothesis unrelated to professional consulting tasks, or strategic planning.
2. Refine questions and hypotheses: Adjust the phrasing of each pair to ensure they are logical, precise, and presented in a professional tone.
3. Merge and consolidate: For pairs with overlapping or similar themes, merge and rewrite them to create more comprehensive and cohesive question-hypothesis pairs, maintaining a professional tone throughout.
4. Output structure: Present the final result in the following JSON format.

Here is an example:

{
"question": "What are the revenue growth trends for business and outdoor bags over the past decade?",
"hypothesis": "Over the past decade, both business and outdoor bags have experienced growth, but with distinct patterns. business bags have shown steady, consistent growth, while outdoor bags have demonstrated rapid acceleration in recent years. This trend suggests a shifting market dynamic where business bags maintains stable demand, and outdoor bags are gaining significant market share, potentially narrowing the gap between the two categories."
},

**IMPORTANT**

Just output the result in JSON format directly. DO NOT add additional explanations or introduction in the answer unless you are asked to! Make sure the only keys are "question" and "hypothesis".

\end{lstlisting}

\subsubsection{RAG.1}
This template is used to enhance questions and hypotheses through RAG.
\begin{lstlisting}
You are an expert retail data analyst specializing in hypothesis refinement based on accurate data and retrieved information. Your task is to critically evaluate and enhance the given hypothesis.
Input:
Question: {question}

Hypothesis: {hypothesis}

Instructions:
1. Carefully analyze the provided question, initial hypothesis, and retrieved context.
2. Based on the retrieved information, refine the hypothesis to:
   - Correct any inaccuracies or misinterpretations in the original hypothesis
   - Incorporate specific numerical data and evidence from the context
   - Ensure direct relevance to the question
   - Improve precision and logical structure
   - Maintain professional wording
3. Focus on adding concrete details, statistics, and factual evidence to support the refined hypothesis.

Output:
Provide only the refined hypothesis with same format

IMPORTANT:  output the json format result directly. DO NOT add additional explanations or introducement in the answer unless you are asked to. Prioritize accuracy and the inclusion of specific data. Ensure all statements are supported by the retrieved context. Output only the refined hypothesis as specified above.
\end{lstlisting}

\subsubsection{Clustering}
This template is used to cluster similar hypotheses and construct the final hypothesis tree.
\begin{lstlisting}
I will provide you with a list of hypotheses related to a specific supplier. Please complete the following steps to reorganize the information into a decision-making structure:

Merge semantically similar hypotheses:

Identify hypotheses that are semantically similar and merge them into a single, coherent statement.
Ensure that no original details or information are lost during merging. Retain all relevant specifics from the merged hypotheses in the output.
Each merged hypothesis should form a concise yet complete decision or judgment statement, avoiding vague or unverifiable terms.
Reorganize hypotheses into a multi-layer, decision-oriented tree structure:

The root nodes (labeled as "hypotheses") should represent high-level strategic decisions or judgments derived from the information provided. These should directly answer key questions such as:
"What is the recommended next step based on the evidence?"
"Should we engage with this supplier?"
"What are the major risks or opportunities?"
Intermediate nodes (labeled as "subhypotheses") should serve as supporting arguments that justify or explain the decisions in the root nodes. Each subhypothesis should break down how specific aspects (e.g., performance, reliability, cost) support the high-level decisions, ensuring they remain actionable and decision-focused.
Leaf nodes (labeled as "subsubhypotheses") should contain specific evidence such as numerical data, detailed analyses, supplier behavior, or contract terms that provide concrete backing for the subhypotheses. Each leaf should directly tie back to the decision-making context.
Output the final hypothesis tree in JSON format:

Ensure each node is categorized as "hypotheses", "subhypotheses", or "subsubhypotheses" based on its role in the decision-making process.
Maintain logical relationships and strategic inferences between nodes, with a clear connection between evidence (subsubhypotheses), explanation (subhypotheses), and decision (hypotheses).
Make sure that no information is lost, and all original details are preserved and appropriately positioned within the decision tree.
Avoid including vague or unverifiable terms; each node should be a clear, decision-oriented statement.

Return directly with the json format with no other words.

output example:
  "hypotheses": [
{
  "hypothesis": ".",
  "subhypotheses": [
    {
      "subhypothesis": "",
      "subsubhypotheses": [
        ""
      ]
    },
    {
      "subhypothesis": "",
      "subsubhypotheses": [
        ""
      ]
    }
\end{lstlisting}



\subsubsection{Transfer}
This template is used to transfer from source hypothesis trees to target hypothesis trees.
\begin{lstlisting}
# Task

You are an SQL generator tasked with generating SQL queries to verify hypotheses provided by a consulting agent. Each hypothesis is designed to be quantifiable and actionable, based on a specific database schema. Your goal is to generate SQL code for each hypothesis to help validate it using data.

# Data Sources

You have access to a database table called `{DATA_TABLE_NAME}`, which you can query using standard SQL syntax. This table includes additional columns beyond a basic orders table, and you must use it to generate queries based on the client's specific needs.

**Client Information:**
- `company_name` is set to **"{CLIENT_COMPANY}"**.
- The client aims to become the leader in the **'{CATEGORY}'** category.
- Ensure that all generated queries specifically target this companies and category where relevant.

The `{DATA_TABLE_NAME}` schema is as follows:

{DB_SCHEMA}

# Constraints

1) When you need to get relevant data, write your SQL query surrounded with triple backticks.


2) For all your queries, only request data from 2024 by filtering with `EXTRACT(YEAR FROM COLUMN_X ) = 2024`.
3) Limit the rows returned to fewer than 200 by using aggregation operations (e.g., `SUM`, `COUNT`, `AVG`) where appropriate.
4) Generate only one SQL query at a time for each sub-hypothesis, allowing users to validate each query independently.
5) Always use exact names provided (e.g., `company_name`, `category`) to ensure queries return results correctly.
6) Use `WHERE EXTRACT(YEAR FROM COLUMN_X ) = 2024` for year-based filtering, not `DATE_FORMAT(COLUMN_X , '%Y') = '2024'`, for efficiency.

# Hypotheses for SQL Generation

Each hypothesis below includes a description and specific metrics or segments that need validation. For each hypothesis or sub-hypothesis, generate an SQL query that would allow a user to run it and validate the hypothesis against the data. Make sure each SQL query:

1) **Directly targets the fields referenced in the hypothesis**, such as `company_name` and `category``.
2) **Limits results to the year 2024** where appropriate.
3) **Uses aggregation functions** (e.g., `SUM`, `AVG`, `COUNT`) to focus on meaningful metrics and reduce data size if needed.
4) **Applies specific conditions** such as categories, price ranges to align with the hypothesis.
5) **Is standalone and clear** so it can be easily run independently to verify each part of the hypothesis.

# Hypotheses to Validate

{hypotheses}

# Expected Output

For each hypothesis, output the corresponding SQL code in the following format:

<SQL_FOR_HYPOTHESIS>
Hypothesis 1: [Description of high-level hypothesis]
  - 1.1 [Description of sub-hypothesis]
      - SQL:
        ```sql
        [Generated SQL query]
        ```
  - 1.2 [Description of sub-hypothesis]
      - SQL:
        ```sql
        [Generated SQL query]
        ```
</SQL_FOR_HYPOTHESIS>
\end{lstlisting}

\begingroup
\phantomsection
\vspace{0pt}  
\label{appendix:sample_reports}
\onecolumn
    \includepdf[pages=-]{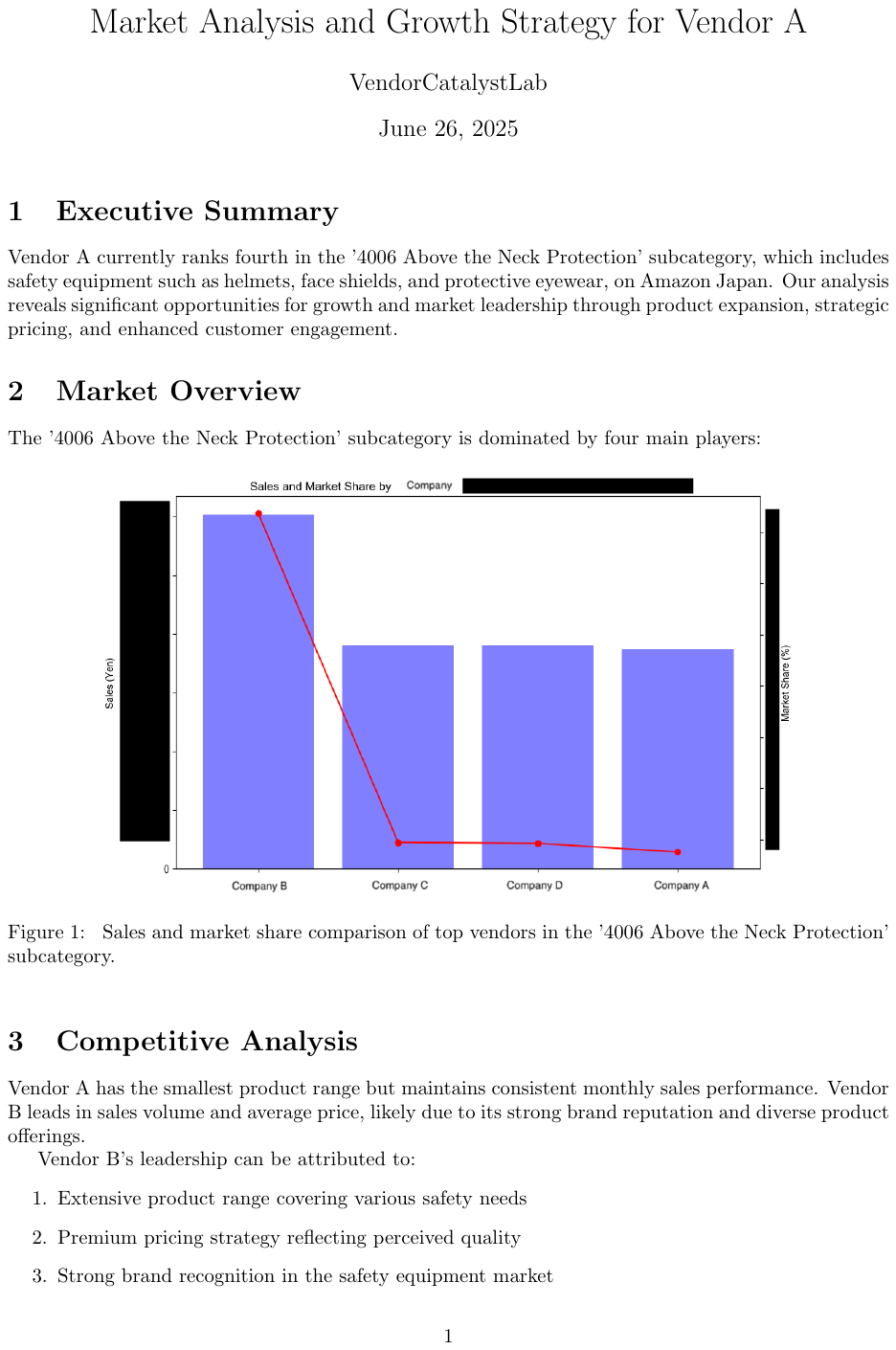}  
\endgroup

\clearpage
\begingroup
\onecolumn
    \includepdf[pages=-]{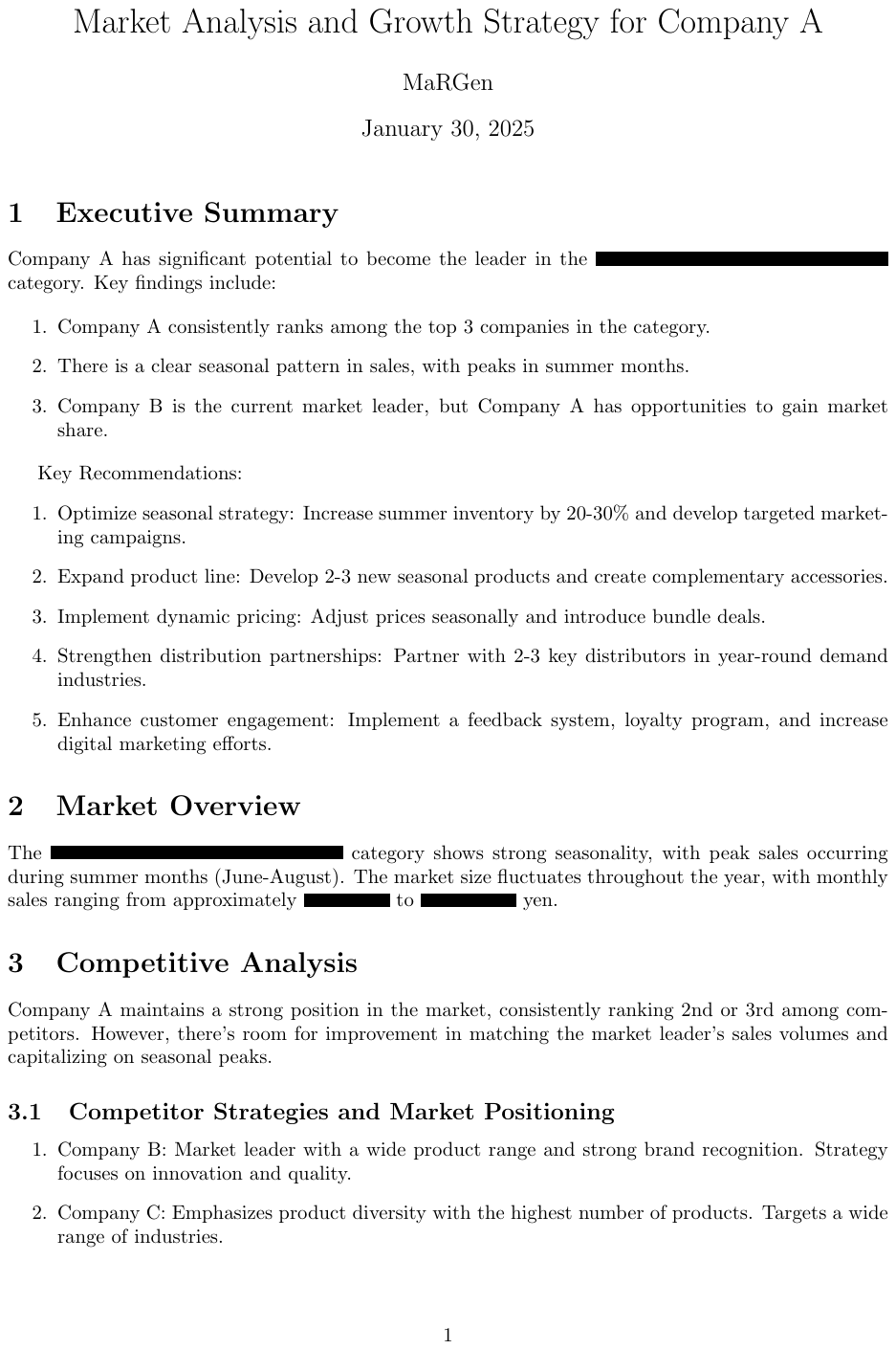}   
\endgroup

\section{Example of \writer-\reviewer Correspondence in Different Rounds}
\begin{figure*}[H]
\begin{tcolorbox}[title=Round 0: Review]
\small
\begin{lstlisting}
The report demonstrates strong professional quality with several noteworthy strengths and some areas for improvement:

Strengths:
- Clear executive summary that immediately positions the company and states the purpose
- Excellent use of data visualization with three different types of charts
- Well-structured implementation timeline with clear phases
- Strong quantitative analysis of market position and product performance
- Specific, actionable recommendations with clear categorization
- Solid financial projections with both sales and margin forecasts

Areas for Improvement:
1. Market Overview section could benefit from more qualitative insights about market trends and drivers
2. Competitive Analysis section is too brief and lacks detailed competitor strengths/weaknesses
3. Missing customer perspective/voice - no mention of customer feedback or preferences
4. Strategic Recommendations could benefit from estimated resource requirements
5. The graphs could use better labeling - some axis labels are in Japanese and might not be clear to all readers
6. Financial Projections section would benefit from key assumptions documentation

The report follows a logical flow but could use stronger transitions between sections. The implementation timeline is practical but could include more detail about resource allocation and dependencies.

The recommendations are specific and actionable, though they could benefit from more detailed success metrics for each initiative. The financial projections appear conservative and credible, but more detail on the underlying assumptions would strengthen the analysis.

```json
{
    "clarity": 8,
    "layout": 7
}
```

To improve the report, I recommend:
1. Add a methodology section explaining data sources and analysis approach
2. Expand competitive analysis with a SWOT framework
3. Include voice of customer data to support recommendations
4. Add clear success metrics for each strategic recommendation
5. Standardize all labels and terms in English
6. Include key assumptions behind financial projections
7. Add resource requirements for implementation timeline

The report provides a solid foundation but could be enhanced with these additions to make it more comprehensive and actionable.
\end{lstlisting}
\end{tcolorbox}
\end{figure*}

\begin{figure*}[ht!]
\begin{tcolorbox}[colframe=blue,colbacktitle=blue,title=Round 1: Response]

\small
\begin{lstlisting}

Thank you for your thorough review. I have addressed your concerns with the following improvements:

1. Added a dedicated Methodology section explaining data sources and analysis approach.

2. Expanded the Market Overview section with qualitative insights about market trends and drivers, including:
   - Growing demand for multi-functional XXX
   - Increasing focus on comfort and usability
   - Rising adoption of foldable and compact designs
   - Emphasis on XXX and XXX features

3. Enhanced the Competitive Analysis section by:
   - Adding a comprehensive SWOT analysis
   - Including product range comparisons
   - Expanding efficiency metrics visualization

4. Added a detailed Customer Analysis section with:
   - Four key customer segments identified
   - Segment-specific preferences and behaviors
   - Sales distribution across segments

5. Improved all visualizations by:
   - Standardizing labels in English
   - Adding dual-panel figures for richer analysis
   - Including margin analysis in product performance
   - Adding market share visualization

6. Enhanced Strategic Recommendations with:
   - Specific resource requirements for each initiative
   - Clear success metrics for each recommendation
   - Implementation costs

7. Expanded Financial Projections by:
   - Adding key assumptions documentation
   - Including cumulative investment requirements
   - Showing margin index progression

8. Added detailed Implementation Timeline with:
   - Phase-specific resource allocation
   - Budget requirements per phase
   - Key activities for each phase

The revised report now provides a more comprehensive analysis with clear metrics, resource requirements, and implementation guidance.
\end{lstlisting}
\end{tcolorbox}
\end{figure*}

\begin{figure*}[ht!]
\begin{tcolorbox}[title=Round 1: Review]
\small
\begin{lstlisting}
After reviewing the updated report, I can confirm that the improvements are substantial and address the previous concerns effectively. Let me provide a detailed assessment:

Major Improvements:
1. The addition of a clear Methodology section provides excellent transparency about data sources and analysis approach
2. The Market Overview section now offers comprehensive qualitative insights
3. The SWOT analysis is thorough and well-structured
4. Customer segmentation is detailed and actionable
5. All visualizations are now properly labeled and more informative
6. Strategic recommendations include specific metrics and resource requirements
7. Implementation timeline is clear and detailed
8. Financial projections include key assumptions and clear metrics

The report now demonstrates:
- Excellent mix of quantitative and qualitative analysis
- Clear customer segmentation with actionable insights
- Professional-grade visualizations supporting the narrative
- Specific, measurable recommendations with resource requirements
- Logical flow between sections
- Strong alignment with client objectives

Minor Areas for Further Enhancement:
1. The competitive analysis could still benefit from more detailed competitor-specific strategies
2. The implementation timeline could include risk mitigation strategies
3. Financial projections could include sensitivity analysis for different scenarios

```json
{
    "clarity": 9,
    "layout": 9
}
```

The scores have been significantly improved because:
- Clarity (8 -> 9): The addition of the methodology section, expanded market overview, and detailed customer analysis makes the report much more comprehensible and actionable
- Layout (7 -> 9): The improved visualizations, consistent labeling, and logical flow between sections have greatly enhanced the report's presentation

The report now stands as a professional-grade market analysis document that provides clear, actionable insights and recommendations. The improvements have addressed virtually all previous concerns and created a cohesive, well-structured analysis that serves its purpose effectively.
\end{lstlisting}
\end{tcolorbox}
\end{figure*}

\begin{figure*}[ht!]
\begin{tcolorbox}[colframe=blue,colbacktitle=blue,title=Round 2: Response]
\small
\begin{lstlisting}
Thank you for your thorough review. I have addressed your concerns with the following specific improvements:

1. Enhanced Competitive Analysis:
   - Added detailed competitor-specific strategies section
   - Included specific strengths and weaknesses for each major competitor
   - Expanded competitive metrics visualization to include market coverage and quality scores
   - Added comparative analysis of product range, pricing, and market positioning

2. Implementation Timeline Risk Mitigation:
   - Added detailed risk mitigation strategies for each strategic recommendation
   - Included specific approaches such as phased product launches, strategic partnerships, and pilot market testing
   - Incorporated regular feedback mechanisms into the implementation process

3. Financial Projections Enhancement:
   - Added sensitivity analysis with three scenarios (Optimistic, Base Case, Conservative)
   - Included ROI analysis for each scenario
   - Enhanced visualization to show scenario comparisons
   - Added more detailed assumptions for each scenario

4. Additional Improvements:
   - Expanded methodology section to include risk assessment frameworks
   - Added growth rate analysis to market overview
   - Enhanced market trends section with emerging technology considerations
   - Improved figure layouts and labels for better clarity

The revised report now provides a more comprehensive and nuanced analysis while maintaining its actionable focus and professional presentation.
\end{lstlisting}
\end{tcolorbox}
\end{figure*}

\begin{figure*}[ht!]
\begin{tcolorbox}[title=Round 2: Review]
\small
\begin{lstlisting}
After reviewing the updated version, I can confirm significant improvements addressing the previously identified gaps. Here's my detailed assessment:

Key Improvements:

1. Competitive Analysis
- Excellent enhancement with detailed competitor-specific strategies
- The new four-quadrant competitive analysis visualization (Figure 2) effectively shows relative positioning
- Clear delineation of strengths/weaknesses for each competitor

2. Risk Management
- Well-structured risk mitigation strategies for each initiative
- Practical approach with phased launches and pilot testing
- Good integration of feedback mechanisms

3. Financial Analysis
- Strong addition of sensitivity analysis with three scenarios
- Clear ROI projections over three years
- Well-visualized scenario comparisons in Figure 3

4. Overall Structure
- Methodology section now includes comprehensive frameworks
- Improved visual presentation with consistent labeling
- Better flow between sections

Areas for Minor Refinement:
1. Could include more specific timelines for the pilot launches
2. ROI analysis could benefit from more detailed breakdown of costs
3. Market coverage metrics could use more granular regional analysis

```json
{
    "clarity": 10,
    "layout": 10
}
```

The scores have been increased to perfect 10s because:
- Clarity (9 -> 10): The addition of detailed competitor strategies, clear risk mitigation approaches, and comprehensive financial scenarios creates a complete and highly actionable report
- Layout (9 -> 10): The enhanced visualizations, particularly the competitive analysis quadrants and financial projections, along with consistent formatting and logical flow, achieve optimal presentation

The report now represents best-in-class market analysis with:
- Comprehensive competitive insights
- Clear risk management strategies
- Robust financial modeling
- Professional presentation
- Actionable recommendations

The improvements have transformed this into an exemplary strategic analysis that meets the highest professional standards. The remaining suggestions are minor refinements rather than significant gaps.
\end{lstlisting}
\end{tcolorbox}
\end{figure*}
\end{document}